\PassOptionsToPackage{table}{xcolor}
\documentclass[10pt,twocolumn,letterpaper]{article}
\usepackage[pagenumbers]{cvpr} %
\usepackage[dvipsnames]{xcolor}

\usepackage{algorithm}
\usepackage{algpseudocode}
\usepackage{adjustbox}
\usepackage{graphicx}
\usepackage{float}
\usepackage{soul}
\usepackage{esvect}
\usepackage{multirow}
\usepackage{pifont}
\usepackage[table]{xcolor}
\usepackage{makecell}

\newcommand{\RNum}[1]{\uppercase\expandafter{\romannumeral #1\relax}}

\newcommand{\cmark}{\ding{51}}%
\newcommand{\xmark}{\ding{55}}%

\newcommand{\methodname}{\textit{EgoGen}}

\definecolor{cvprblue}{rgb}{0.21,0.49,0.74}
\usepackage[accsupp]{axessibility}
\usepackage[pagebackref,breaklinks,colorlinks,citecolor=cvprblue,hypertexnames=false]{hyperref}

\def\paperID{7475} %
\def\confName{CVPR}
\def\confYear{2024}

\title{EgoGen: An Egocentric Synthetic Data Generator}

\author{Gen Li$^{1}$ \quad Kaifeng Zhao$^{1}$ \quad Siwei Zhang$^{1}$ \quad Xiaozhong Lyu$^{1}$ \\ \quad Mihai Dusmanu$^{2}$ \quad Yan Zhang$^{1}$ \quad Marc Pollefeys$^{1,2}$ \quad Siyu Tang$^{1}$ \vspace{0.3em} \\
{\normalsize $^1$ETH Z\"urich} \quad
{\normalsize $^2$Microsoft}\\
{\normalsize \url{https://ego-gen.github.io/}}
\vspace{-5mm}
}

\begin{document}

\twocolumn[{%
\renewcommand\twocolumn[1][]{#1}%
\maketitle
\begin{center}
    \centering
    \captionsetup{type=figure}
    \includegraphics[width=\textwidth]{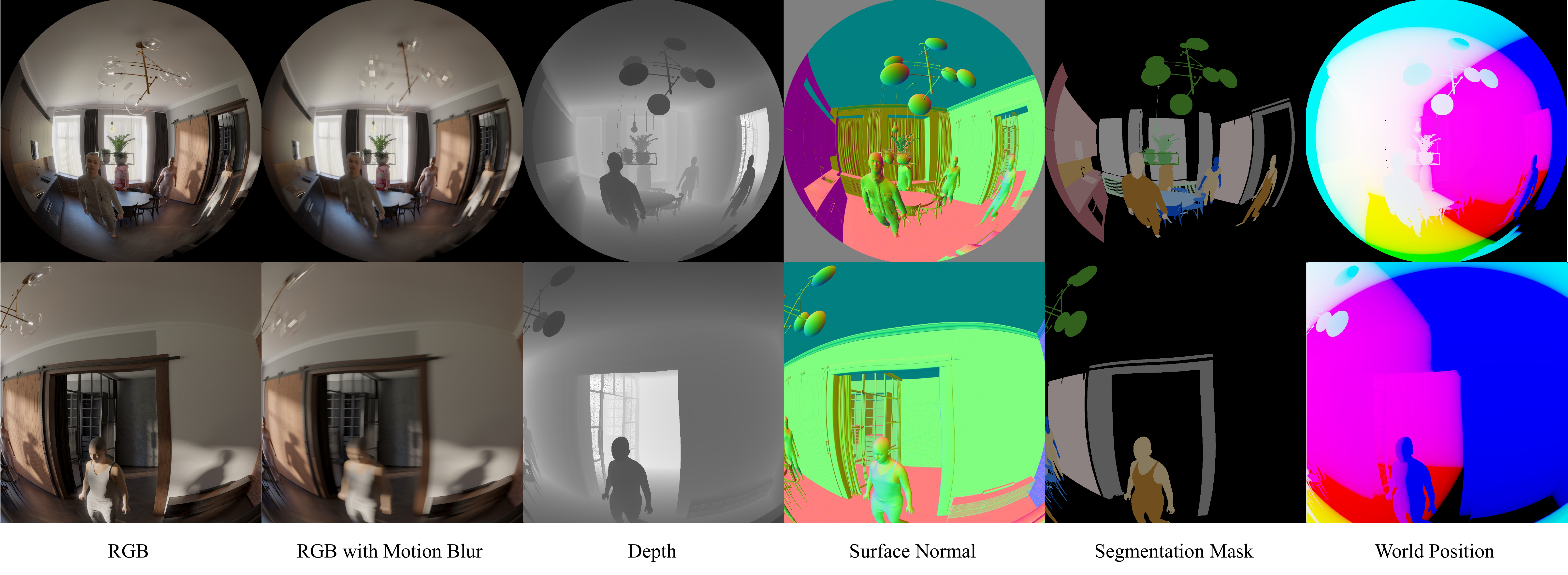}
    \captionof{figure}{EgoGen: a scalable synthetic data generation system for egocentric perception tasks, with rich multi-modal data and accurate annotations. We simulate camera rigs for head-mounted devices (HMDs) and render from the perspective of the camera wearer with various sensors. Top to bottom: middle and right camera sensors in the rig. Left to right: photo-realistic RGB image, RGB with simulated motion blur, depth map, surface normal, segmentation mask, and world position for fisheye cameras widely used in HMDs.}
    \label{fig:teaser}
\end{center}%
}]

\begin{abstract}
Understanding the world in first-person view is fundamental in Augmented Reality (AR).
This immersive perspective brings dramatic visual changes and unique challenges compared to third-person views. %
Synthetic data has empowered third-person-view vision models, but its application to embodied egocentric perception tasks remains largely unexplored.
A critical challenge lies in simulating natural human movements and behaviors that effectively steer the embodied cameras to capture a faithful egocentric representation of the 3D world. 
To address this challenge, we introduce \methodname, a new synthetic data generator that can produce accurate and rich ground-truth training data for egocentric perception tasks. 
At the heart of \methodname~is a novel human motion synthesis model that directly leverages egocentric visual inputs of a virtual human to sense the 3D environment. Combined with collision-avoiding motion primitives and a two-stage reinforcement learning approach, our motion synthesis model offers a closed-loop solution where the embodied perception and movement of the virtual human are seamlessly coupled. 
Compared to previous works, our model eliminates the need for a pre-defined global path, and is directly applicable to dynamic environments. 
Combined with our easy-to-use and scalable data generation pipeline, we demonstrate  \methodname’s efficacy in three tasks: mapping and localization for head-mounted cameras, egocentric camera tracking, and human mesh recovery from egocentric views. \methodname~will be fully open-sourced, offering a practical solution for creating realistic egocentric training data and aiming to serve as a useful tool for egocentric computer vision research.

\end{abstract}    
\vspace{-7pt}
\section{Introduction}
The analysis of visual input from front-facing egocentric cameras is crucial for applications that benefit from a first-person perspective, mirroring the natural human experience~\cite{Ego4D2022CVPR, egoego, Zhang:ECCV:2022}. AR devices, for instance, can utilize this viewpoint to amplify user immersion. Such cameras can also cater to individual preferences, providing custom visual assistance for those with impaired vision \cite{zhao2019designing,fiannaca2014headlock}.

Despite its potential, egocentric perception faces challenges, primarily due to the scarcity of labeled data. Although datasets like Ego4D~\cite{Ego4D2022CVPR}, ADT~\cite{DBLP:journals/corr/abs-2306-06362}, Epic-Kitchen~\cite{Damen2021PAMI} and HoloAssist~\cite{HoloAssist2023} exist, creating such datasets with rich and accurate annotations is costly and raises privacy concerns~\cite{Zhang:ECCV:2022}. 
Alternatively, using graphics techniques to render synthetic multi-modal visual data has proven to be cost-effective and successful in training deep learning models, such as 3D human body estimation~\cite{Black:CVPR:2023} and facial landmark detection~\cite{wood2021fake}.

Creating egocentric synthetic data is challenging because egocentric cameras capture the complex interplay of body movements and the environment from the camera wearer's viewpoint.
Modeling the intricate details and variations in human behavior presents a significant challenge.

To tackle this problem, we introduce \methodname, an egocentric synthetic data generation approach that simulates data from embodied sensors, i.e., front-facing cameras in head-mounted devices (HMD). 
While the ultimate goal is to simulate human behaviors that are indistinguishable from reality, in this work, we focus on creating virtual humans (i.e., camera wearers) that can explore and avoid obstacles in the 3D world that is not only complex and dynamic but could potentially include other {\it moving} virtual humans. 

Specifically, we propose a novel generative human motion model.
Our key insight is that body movement and embodied perception should be seamlessly coupled.
As Gibson stated, {\it ``We must perceive in order to move, but we must also move in order to perceive''}, our egocentric perception is crucial for identifying obstacles, navigating an environment, and planning actions. Our body movements are not solely a response to visual stimuli; they also change our egocentric perception.
Therefore, the key idea of our motion model is to enable virtual humans to {\it see} their environment with {\it egocentric} visual inputs and respond accordingly by learning a policy to control a set of collision-avoiding motion primitives (CAMPs) composable for synthesizing long-term, diverse human motions. 
Due to the unbounded and high-dimensional latent action space of our generative motion primitive model, direct policy training through rendered egocentric images is often unstable \cite{zhuang2023robot}. 
Therefore, we propose a two-stage reinforcement learning scheme using an efficient {\it egocentric} visual proxy to couple egocentric visual cues and body movements seamlessly. 
In addition, we use an ``attention'' reward to incentivize egocentric perception behaviors, i.e.,~looking in the desired direction.

Empirical results showcase the benefits of our egocentric perception-driven motion framework, which does not require a pre-calculated walking path in 3D scenes as in~\cite{hassan_samp_2021, zhao_synthesizing_2023, DBLP:journals/corr/abs-2304-02061}. Instead, it empowers virtual humans to perceive the environment from their own viewpoint, enabling them to navigate, circumvent obstacles, and plan movements to reach the destination. 
Moreover, our model generalizes well to dynamic environments, even with training limited to static settings.
By training virtual humans independently using CAMPs, our method synthesizes emergent multi-human behaviors without relying on multi-agent reinforcement learning algorithms. 
Egocentric visual cues are essential to build exploratory and generalizable motion models that unify navigational planning and movement control in complex and dynamic environments.

Building upon CAMPs, we further create a scalable data generation pipeline for \methodname~that outfits virtual humans with clothing, automates cloth animation, and integrates 3D assets from various sources. 
We validate \methodname's efficacy across three egocentric perception tasks. The high-quality synthetic data with precise ground truth annotations consistently improve the performance of state-of-the-art methods. In summary, the contributions of this work are:
\begin{enumerate}
    \item We introduce \methodname, a generative and scalable synthetic data generation approach specifically tailored for egocentric perception tasks. 
    \item We introduce novel motion primitives based on egocentric visual cues, 
    enabling diverse and realistic human motion synthesis in 3D scenes. 
    These primitives empower virtual humans to handle complex scenarios, such as dynamic environments and crowd motion without relying on multi-agent reinforcement learning. 
    \item \methodname~enables us to augment existing real-world egocentric datasets with synthetic images. Quantitative results demonstrate enhanced performance in state-of-the-art algorithms on mapping and localization for head-mounted cameras, egocentric camera tracking, and human mesh recovery from egocentric views.
\end{enumerate}

\section{Related Work}

\textbf{Human-Related Simulators and Synthetic Data.}
Previous works primarily focus on simulating robots \cite{quigley2009ros, DBLP:conf/iros/MontemerloRT03, szot2021habitat, 6386109, savva2019habitat} and autonomous cars \cite{dosovitskiy2017carla, nvidiadrivesim, baltodano2015rrads, rong2020lgsvl}. While some incorporated animated digital humans, like in \cite{nvidiadrivesim, hakada2022unrealego, DBLP:journals/corr/abs-2310-13724}, these efforts relied on pre-recorded motion sequences. Rendering images of people to train perception models has been widely studied such as~\cite{egoego, 2398356.2398381,STRAPS2020BMVC,ebadi2022psphdri,pumarola20193dpeople, fabbri2018learning,hakada2022unrealego}. 
In particular, \cite{tome2019xr} 
offers large-scale synthetic data for egocentric camera wearer pose estimation but relies on mocap data, lacking realistic and spontaneous interactions with the digital world.
In contrast, \methodname~closes the gap in egocentric synthetic data generation for head-mounted devices. 
Please refer to \cref{sec:srelated_work} for detailed comparisons.

\noindent\textbf{Human Motion Synthesis.}
Generating high-fidelity human motions and interactions with 3D scenes is widely studied in graphics \cite{kovar2008motion, clavet2016motion, holden2020learned, starke_deepphase_2022, starke_neural_2019, holden_phase-functioned_2017}. 
While they can generate high-quality motion, it's usually deterministic.
Synthesizing physically plausible human motions has been extensively studied~\cite{coumans2021, todorov2012mujoco, peng2021amp, PhysicsVAE, hassan2023synthesizing, rempe2023trace}. However, they struggle with generalization to different body shapes. 
For example, \cite{rempe2023trace} explicitly created 2048 humanoids to improve body shape generalizability.
Time series models~\cite{petrovich2021actionconditioned, tevet2022human, zhang2022motiondiffuse} synthesize the stochastic motions of diverse people well.
However, in~\cite{tevet2022human,zhang2022motiondiffuse}, their generated motion sequences have limited lengths and human-scene interactions are not explicitly considered. 
Autoregressive methods~\cite{ling2020character,rempe2021humor,zhang2022wanderings,zhao_synthesizing_2023} can produce perpetual motions. 
In particular, \cite{zhang2022wanderings} can generalize to diverse body shapes and synthesize long-term human motions.

Our egocentric perception-driven motion synthesis model is closely related to \cite{zhang2022wanderings, zhao_synthesizing_2023}, but distinguishes itself w.r.t.:
(1) Enabling virtual humans to explore using egocentric visual cues, without predefined paths.
(2) Synthesizing egocentric perception behaviors beyond locomotion, e.g., looking in certain directions.
(3) Handling dynamic environments and multi-agent behavior without re-training. %

\noindent\textbf{Mapping and Localization for AR.}
Localization and mapping from images is a long-standing problem known as:
Photogrammetry~\cite{haala2012performance, abdel2012efficient}; Structure-from-Motion~(SfM)~\cite{Govindu2001Combining, Pollefeys2004Visual, Wilson2014Robust, Schoenberger2016Structure, Snavely2006Photo}; Simultaneous Localization and Mapping~(SLAM)~\cite{Davison2003Real, Nister2004Visual, Mouragnon2006Real, Klein2007Parallel}.
Researchers have worked to make SLAM amenable for edge hand-held or head-mounted devices~\cite{Klein2007Parallel, ARKit, ARCore}.
Cloud-based services like Google's Visual Positioning System~\cite{Google2019Google}, Niantic's Lightship~\cite{Niantic2021Niantic}, and Microsoft's Azure Spatial Anchors~\cite{Microsoft2019Announcing} have made visual localization and mapping more accessible.
Benchmarking efforts have arisen for small-scale AR scenarios~\cite{shotton2013scene, kendall2015posenet}, touristic landmarks~\cite{Schonberger2017Comparative, Jin2021Image}, and large-scale AR-device based localization~\cite{Sattler2017Benchmarking, sarlin2022lamar} to evaluate these systems.

\noindent\textbf{Egocentric Human Pose Estimation.}
Estimating 3D bodies from RGB images is widely studied from third-person views~\cite{kanazawa2018end, kolotouros2019cmr, kolotouros2019learning, SMPL-X:2019, Kocabas_PARE_2021, li2022cliff, Bogo:ECCV:2016, kolotouros2021probabilistic}, and egocentric views~\cite{luo2020kinematics,yuan2019ego,shiratori2011motion,tome2020selfpose,tome2019xr,guzov2021human, ng2020you2me, liu20214d, ye2023decoupling, zhang2023probabilistic, wang2023scene}, mostly requiring expensive real-world data paired with ground truth annotations.
Besides RGB images, depth images offer explicit 3D information, mitigating scale and shape ambiguity, with the potential to enable broader AR/VR applications. 
However, depth-based methods, especially for the egocentric view, are underexplored.
Most existing works~\cite{moon2018v2v, xiong2019a2j, martinez2020residual, haque2016towards, shotton2012efficient, ye2011accurate, wang2016human, rafi2015semantic} predict 3D body skeletons without expressive body meshes, struggling with challenges like severe body truncations and scene occlusions typical in egocentric views.
Such limited attention mainly stems from the scarcity of data, as obtaining high-quality human mesh annotations for real-world depth images is labor-intensive.

\section{Ego-Sensing Driven Motion Synthesis}
\label{sec:3}
\begin{figure*}[t]
    \centering
    \includegraphics[width=\linewidth]{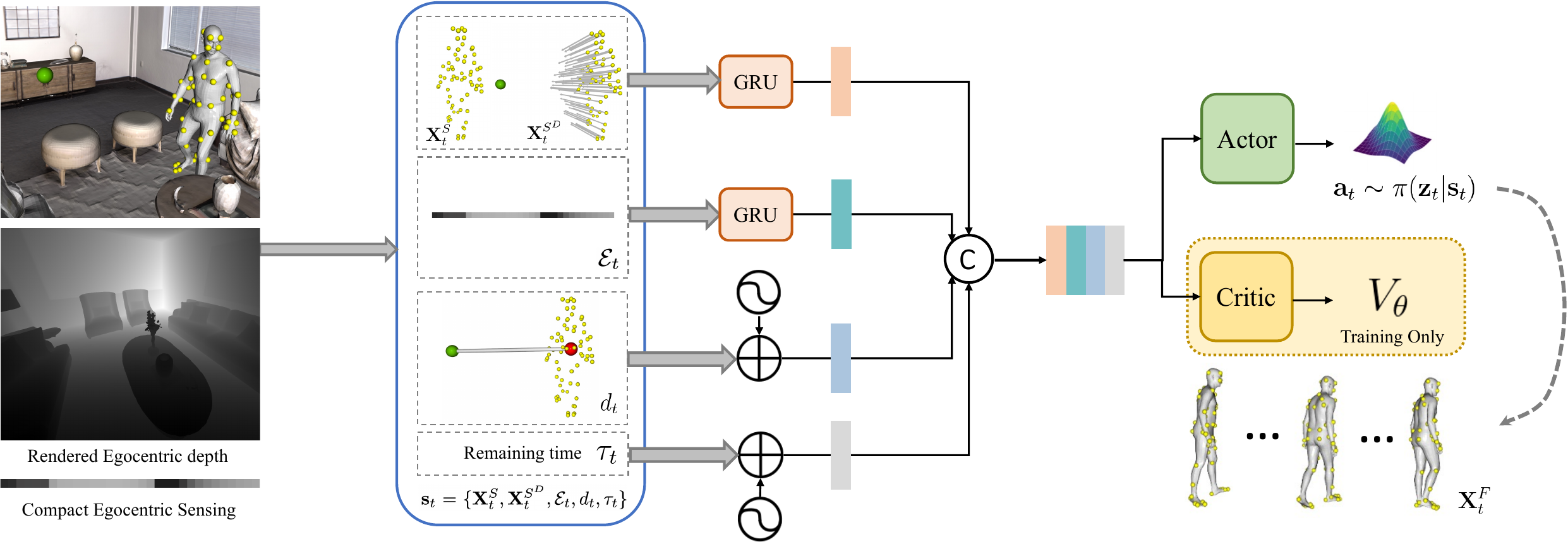}
    \caption{Policy network architecture. We learn a generalizable mapping from motion seed body markers $\mathbf{X}_t^S$, marker directions $\mathbf{X}_t^{S^D}$, egocentric sensing $\mathcal{E}_t$, and distance to the target $d_t$ to CAMPs. The policy learns a stochastic collision avoiding action space to predict future body markers $\mathbf{X}_t^F$. For illustration purposes, we visualize only one frame of $\mathbf{X}_t^S$ and $\mathcal{E}_t$. See Sec.~\ref{sec:3-1} and \ref{sec:3-2} for details.} 
\label{fig:method-overview}
\vspace{-2mm}
\end{figure*}

To close the loop for the interdependence between egocentric synthetic image data and human motion synthesis, we use deep reinforcement learning (RL), integrating egocentric vision cues to synthesize human motions as described in Sec.~\ref{sec:3-1} and \ref{sec:3-2}. Subsequently, we extend learned policies to generate emergent multi-agent behaviors, as in Sec.~\ref{sec:3-3}.

\subsection{Ego-Sensing Driven Motion Primitives}
\label{sec:3-1}

Generating realistic egocentric data requires diverse and lifelike human motion synthesis.
In this work, we consider arguably the most common everyday behaviors: navigating towards goals with egocentric perception while avoiding collisions with obstacles and people in dynamic 3D scenes. 

\noindent\textbf{Overview.} %
Following recent literature \cite{ling2020character, zhang2022wanderings, controlvae, zhao_synthesizing_2023}, we employ deep RL to train control policies on learned latent spaces that characterize natural human motions. 
However, unlike these previous works that only consider simple static scenes, we leverage egocentric perception and propose collision-avoiding motion primitives (CAMPs) to enable virtual humans to self-explore and navigate in a dynamic environment. 
Specifically, CAMPs are trained jointly to produce collision-free motion sequences. %
At each timestep $t$, the agent observes the state $\mathbf{s}_t$, performs an action $\mathbf{a}_t$, and receives a reward $r_t = r(\mathbf{s}_t, \mathbf{a}_t, \mathbf{s}_{t+1})$, where $\mathbf{s}_{t+1}$ represents the next state of the environment due to $\mathbf{a}_t$.

\noindent\textbf{Egocentric Sensing As Depth Proxy.} We aim to sample actions given by a policy to synthesize realistic human motions.
Egocentric perception-driven motion synthesis should arguably use egocentric vision as input. However, depth rendering is costly and RL requires billions of samples to converge \cite{DBLP:conf/corl/AgarwalKMP22, DBLP:conf/iclr/WijmansKMLEPSB20}. Besides, directly training RL with visual data can be unstable \cite{zhuang2023robot}.
We thereby use a cheap-to-compute \textit{egocentric sensing} $\mathcal{E}_t$ as a proxy for depth images as illustrated in Fig.~\ref{fig:method-overview}. $N$ rays are cast evenly from the midpoint of two eyeballs, i.e., the location of the egocentric camera. The field of view $[\theta_{min}, \theta_{max}]$, centered on the 2D projection of the viewing direction $\vv{\mathbf{v}}$, limits the agent's perception to the front area. Rays stop at collisions, with collision detection in 2D. See more details in 
\cref{sec:s2-1}.

\noindent\textbf{Agent Representation.} The agent is a virtual human represented by an SMPL-X mesh \cite{SMPL-X:2019}. We further compact the body representation by selecting $M=67$ body surface markers $\mathbf{x} \in \mathbb{R}^{M\times3}$ on the mesh following \cite{zhang2021mojo}.

\noindent\textbf{Environment.} Inspired by \cite{DBLP:conf/iros/YanLZB20}, we aim to learn a library of composable CAMPs.
We implement a \textit{finite-horizon} environment based on the generative motion primitive model from~\cite{zhang2022wanderings}. 
Specifically, a motion primitive is defined as a 0.5-second motion clip containing $T=20$ frames in the canonical coordinate, and each frame contains a single agent representation.
The primitive model~$\mathcal{P}$ is based on the C-VAE framework~\cite{NIPS2015_8d55a249}, which takes the first $T_s=2$ frames as the condition, and models a conditional probability of the next $T-T_s$ frames.
Compared to \cite{zhang2022wanderings} trained on the AMASS dataset~\cite{mahmood2019amass} with many sport motion sequences, 
we train~$\mathcal{P}$ using the SAMP dataset~\cite{hassan_samp_2021}, which focuses on daily activities, better suited for HMD use cases.
Our \textit{action space} $\mathcal{A}$ is the pretrained 128D latent space of $\mathcal{P}$, and the \textbf{action} $\mathbf{a}_t$ can be randomly sampled from $\mathcal{A}$.

With the input of a random action $\mathbf{a}_t \in \mathbb{R}^{128}$ and a motion \textbf{s}eed $\mathbf{X}^S_t = [\mathbf{x}_t^0, \mathbf{x}_t^{T_s-1}]$ (history frames), $\mathcal{P}$ predicts \textbf{f}uture frames $\mathbf{X}^F_t = [\mathbf{x}_t^{T_s}, ...,\mathbf{x}_t^{T-1}]$ of the current motion primitive $\mathbf{X}_t = [\mathbf{X}^S_t, \mathbf{X}^F_t] \in \mathbb{R}^{T \times M \times 3}$, which represents a short sequence of human motion spanning 0.5~s:
\begin{align*}
    \mathbf{X}^F_t = \mathcal{P}(\mathbf{X}^S_t, \mathbf{a}_t)
\end{align*}

\noindent\textbf{State.} To preserve Markov property \cite{pardo2018time}, the state is defined as $\mathbf{s}_t = \{\mathbf{X}_t^S, \mathbf{X}_t^{S^D}, \mathcal{E}_t, d_t, \tau_t\}$, in which $\mathbf{X}_t^{S^D} \in \mathbb{R}^{T_s \times M \times 3}$ denotes the normalized direction of each marker seed to the target, $\mathcal{E}_t \in \mathbb{R}^{T_s \times N}$ denotes the egocentric sensing depth proxy, $d_t$ denotes the distance from the pelvis to the target, and $\tau_t$ denotes the remaining time. See Fig.~\ref{fig:method-overview}.

\label{att_rew}
\noindent\textbf{Reward.} To synthesize egocentric perception behaviors,
we use an ``attention reward'' to incentivize the virtual human to look in specific directions: $r_{attention}=\cos \langle \vv{\mathbf{v}}, \vv{\mathbf{a}} \rangle$, where $\vv{\mathbf{a}}$ is the attention direction from the head joint to the viewing target. The reward function is defined as:
\begin{equation*}
    r_t = r_{cont.} + r_{dist} + r_{ori} + r_{attention} + r_{pene} + r_{pose} + r_{succ},
\end{equation*}
where $r_{cont.}$ enforces valid foot contact and minimizes foot skating; $r_{dist}$ encourages reaching the target; $r_{ori}$ aligns the body forward direction with the target; $r_{pene}$ guides collision avoidance; $r_{pose}$ reduces unrealistic human poses; and $r_{succ}$ is a sparse reward when reaching the target. 

\noindent\textbf{Episode Termination.} To handle collisions beyond \cite{zhao_synthesizing_2023}, we employ multiple \textit{termination} signals to conclude an episode if the generated motion primitive $\mathbf{X}_t$ satisfies any of the following conditions:
\begin{itemize}
    \item Success: The virtual human reached the target.
    \item Penetration: The virtual human collides with the obstacle.
    \item Timeout: The virtual human did not reach the target within the maximum timesteps.
\end{itemize}

\subsection{Training Collision-Avoiding Stochastic Policies}
\label{sec:3-2}

\noindent\textbf{Algorithm.} We use Proximal Policy Optimization \cite{schulman2017proximal} (PPO) to learn a generalizable mapping from various egocentric sensing and body configurations to CAMPs. 
Instead of extensive manual data collection for all possible input combinations, we leverage the stochastic nature of the PPO policy. Through exploration and sampling actions, the agent traverses the scene and generates varied egocentric sensing and body configurations, diversifying the training data.

Instead of training each CAMP independently for every single step, we use PPO to train a sequence of CAMPs jointly in multi-step collision avoidance tasks. This approach can benefit choosing a more favorable CAMP which makes the subsequent action easier.

\noindent\textbf{Network.} The network architecture is shown in Fig \ref{fig:method-overview}. The actor and critic network share a feature extraction trunk to encode the state $\mathbf{s}_t$: the motion seed ($\mathbf{X}_t^S$ and $\mathbf{X}_t^{S^D}$) and the egocentric sensing $\mathcal{E}_t$ are encoded using RNNs; the rest of scalar states are encoded using positional encoding~\cite{vaswani2017attention}\label{posenc}. The actor predicts a stochastic policy $\mathbf{a}_t \sim \pi(\mathbf{z}_t|\mathbf{s}_t)$ conditioned on the current state $\mathbf{s}_t$, where $\mathbf{z}_t \sim \mathcal{N}(\mathbf{\mu}, \Sigma)$. $\mathbf{\mu}$ and $\Sigma$ are the mean and variance of the learned action space.

\noindent\textbf{Objective Function.} The objective function includes the policy surrogate $L^{CLIP}$, the value function error term $L^{VF}$ to evaluate the value prediction $V_\theta$, and an entropy bonus $L^{S}$ to encourage exploration: %
\begin{align*}
    L = L^{CLIP} + c_1 L^{VF} + c_2 L^{S} %
\end{align*}
where $c_1, c_2$ are coefficients. See more details in \cref{sec:s-ppo}.
\noindent\textbf{RL Pre-training and Finetuning.} Training in crowded scenes, e.g.~Replica \cite{replica19arxiv}, requires additional steps.
Because the action space $\mathcal{A}$
is an unbounded Gaussian, RL exploration while predicting reasonable human poses can be challenging.
We first pretrain the policy with a higher $r_{pene}$ weight without \textit{penetration termination}. After convergence, we finetune it with strict termination constraints using a signed distance field (SDF) for penetration detection. Please refer to 
\cref{sec:s-reward} 
for the formulation and weighting of each reward and training detail.

\subsection{Compositing Learned Motion Primitives}
\label{sec:3-3}

\begin{algorithm}[t]
\caption{Crowd motion synthesis with learned CAMPs\label{alg:cap}}
\begin{algorithmic}
\algrenewcommand\algorithmiccomment[1]{\hfill\(\triangleright\) #1}
    \State \textbf{Result:} Multi-human locomotion w/ collision avoidance;
    \State \textbf{Init:} crowd size $C$, marker seed for each human $\mathbf{X}^S_{c}$; %
    \For{$step \gets 1$ \textbf{to} $max\_steps$} \Comment{env. finite horizon}
        \For{$c \gets 1$ \textbf{to} $C$} \Comment{for each human}
            \State update all locations with $\{bbox(\mathbf{X}^S_{c})\}_{c=1:C}$
            \State compute egocentric sensing $\mathcal{E}_c$; 
            \State execute the action that maximizes the expected return, and produce one CAMP;
        \EndFor
    \EndFor
\end{algorithmic}
\end{algorithm}

Although CAMPs are trained solely with static scenes, their direct application to dynamic settings is achieved by decomposing jointly trained CAMPs into individual motion primitives and re-compositing them. 
Our model demonstrates effective generalization by selecting the next best motion primitive from the learned CAMP library to maximize the expected return, provided that the egocentric sensing is updated with the most recent obstacle location at each timestep.
Furthermore, our model is directly applicable to tasks involving complex interactions with other virtual humans. To synthesize crowd motion (Alg.~\ref{alg:cap}), each virtual human employs the same policy to navigate and avoid others. 
To a specific virtual human, others are seen as dynamic obstacles, represented by body bounding boxes for avoidance.
Acknowledging the inherent delay in human reactions when avoiding dynamic obstacles \cite{Aivar2008}, agents take a single CAMP sequentially, instead of in parallel, i.e. the first agent generates its first CAMP and waits for others to complete their first CAMP before all agents move on to prepare their second CAMP. To ensure successful collision avoidance, the agent's egocentric sensing is updated before taking a new action.
This composition of CAMPs synthesizes emergent multi-human behaviors {\it without} multi-agent RL algorithms (see Sec.~\ref{sec:5-1}), enhancing the generalization and scalability.

\section{Egocentric Synthetic Data Generation}

\begin{figure}[t]
    \centering
    \setlength{\tabcolsep}{0.0130\linewidth}
    \includegraphics[width=1.0\linewidth]{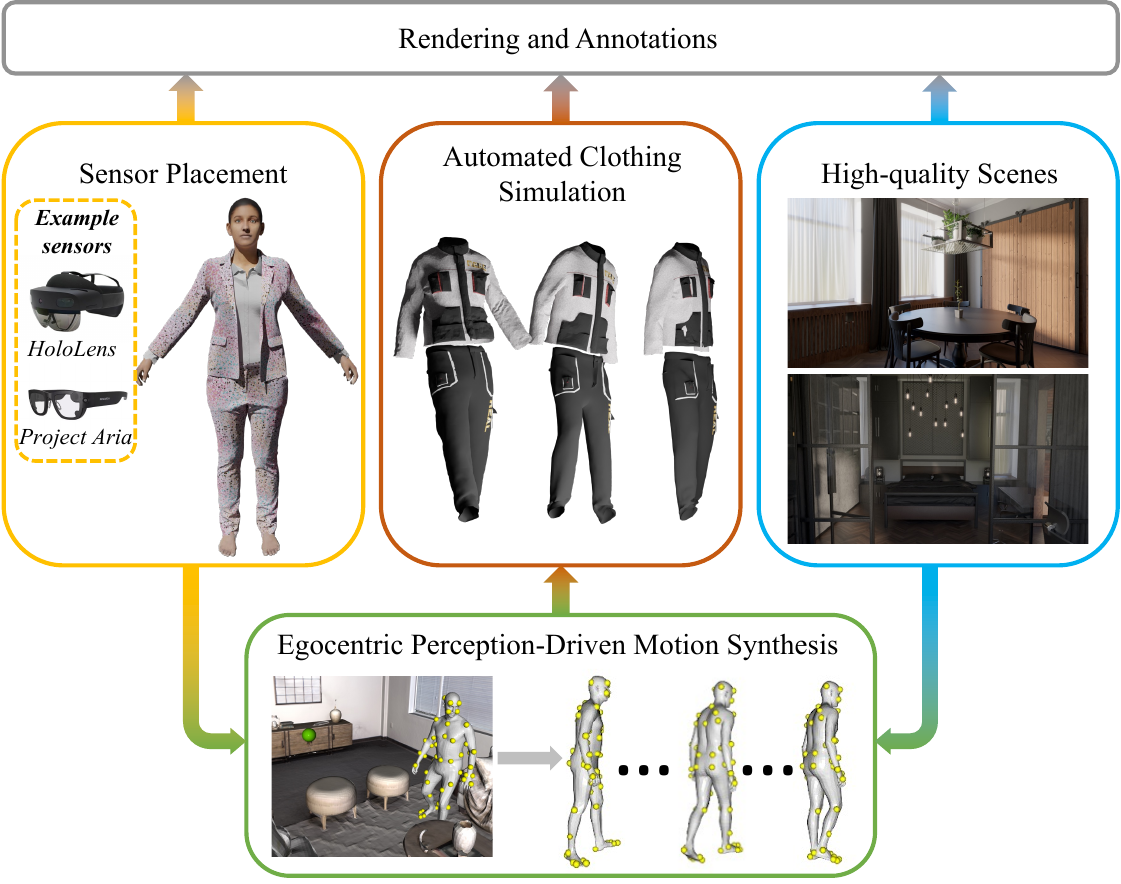}
    \caption{Overview of \textit{EgoGen}. Through generative motion synthesis (Sec.~\ref{sec:3}), %
    we further enhance egocentric data diversity by randomly sampling diverse body textures (ethnicity, gender) and 3D textured clothing through an automated clothing simulation pipeline (Sec.~\ref{sec:clothing}). With high-quality scenes and different egocentric cameras, we can render photorealistic egocentric synthetic data with rich and accurate ground truth annotations (Sec.~\ref{sec:render}). \vspace{-3mm}}
\label{fig:pipeline}
\end{figure}

Synthesizing realistic egocentric perception-driven human motions (as detailed in Sec.~\ref{sec:3}) forms the foundation of simulating egocentric synthetic data.
An overview of our egocentric data generation pipeline \methodname, is shown in Fig.~\ref{fig:pipeline}.

\subsection{Embodied Camera Placement}

\label{sec:cam_place}
Similar to existing AR devices, we use the head pose to define the egocentric viewing direction $\vv{\mathbf{v}}$. %
Our development is based on Blender~\cite{bldddd}.
We use the SMPL-X~\cite{SMPL-X:2019} mesh to position the egocentric camera between the two eyeballs. The camera's viewing direction ($\vv{\mathbf{v}}$) is perpendicular to the plane determined by the two eye bones in the armature.
We also support simulating multi-camera rigs as shown in Fig.~\ref{fig:teaser}.
When the body moves (Sec.~\ref{sec:3-3}), we can synthesize egocentric videos with continuously updated camera poses.

\subsection{Body Texture and Clothing}
\label{sec:clothing}
To enhance \methodname's synthetic data realism, we dress virtual humans using human textures and 3D clothing assets from BEDLAM~\cite{Black:CVPR:2023, Meshcapade}, including 50 male and 50 female skin albedo textures from seven ethnic groups.

Unlike prior works~\cite{Black:CVPR:2023, yang2023synbody} relying on unscalable commercial software for clothing dynamics simulation, we automate it for diverse synthesized motions and body shapes, minimizing manual effort. Each garment mesh is in a consistent rest pose, i.e., A-Pose (See Fig.~\ref{fig:pipeline} middle-left). 
For each motion sequence, we first repose it to match the body pose in the first frame using linear blend skinning. This involves initializing the clothing geometry by sampling pose and shape blend shapes, along with skinning weights from the nearest multiple SMPL-X vertices in A-Pose. Then we simulate upper and lower garments separately using a state-of-the-art clothing simulation network~\cite{grigorev2022hood}.

\subsection{Rendering and Annotations}
\label{sec:render}
\textit{EgoGen} supports simulating diverse head-mounted devices with different camera models, such as fisheye and pinhole cameras.
Given the camera's intrinsic parameters and relative poses within the camera rig, we can simulate AR devices like Project Aria glasses~\cite{aria} and HoloLens~\cite{HoloLens}, facilitating synthetic data generation for real-world applications.
Camera extrinsic is determined by our generative human motion model. We use Blender~\cite{bldddd} to render photorealistic egocentric image sequences with motion blur. We also render out a rich set of ground truth annotations, such as depth maps, surface normals, segmentation masks, world positions, optical flow, etc for egocentric perception tasks.
\section{Experiments}

We assess the motion quality, generalizability, and diversity of our motion model, highlighting its ability to generalize to unseen complex tasks and comparing it with recent baselines (Sec.~\ref{sec:5-1}).
We evaluate our proposed egocentric sensing as a depth proxy for enhancing agent exploration (Sec.~\ref{sec:5-2}) and conduct ablation studies (Sec.~\ref{sec:5-3}).

We further demonstrate the effectiveness of \methodname~on three egocentric computer vision tasks in Sec.~\ref{lamar}, and~\ref{exp-egobody}. By incorporating synthesized egocentric images, we can enhance the performance of the state-of-the-art algorithms.

\subsection{Evaluation of Learned CAMPs}
\label{sec:5-1}

\begin{table}[t]
 \caption{Evaluation of motion synthesis in scenes with moving obstacles, multiple humans, and path diversity. $\downarrow$: lower is better; $\uparrow$: higher is better. The best results in each scenario are in boldface. * denotes an improved version for fair comparison. (Sec.~\ref{sec:5-1})}
\centering
\begin{adjustbox}{width=0.45\textwidth}
 \begin{tabular}{||c c c c c||} 
 \hline
  Evaluation & Metrics & GAMMA*~\cite{zhang2022wanderings} & DIMOS*~\cite{zhao_synthesizing_2023} & Ours \\ 
 \hline\hline
    \multirow{4}{*}{Mov. obs.}    & SR (\%) $\uparrow$ & 96 & 83 & \textbf{100} \\
        & Dist. (m) $\downarrow$ & 0.29 & 0.55 & \textbf{0.06} \\
        & Cont. $\uparrow$ & 0.95 & 0.96 & \textbf{0.97} \\
        & Pene-S. (\%) $\downarrow$ & 9.2 & 8.4 & \textbf{3.4} \\
\hline

\multirow{4}{*}{2 humans}    & SR (\%) $\uparrow$ & 95  & 88 & \textbf{100} \\
        & Dist. (m) $\downarrow$ & 0.32 & 0.41 & \textbf{0.07} \\
        & Cont. $\uparrow$ & 0.96 & \textbf{0.98 }& 0.97 \\
        & Pene-H. $\downarrow$ & 27.6 & 10.7 & \textbf{0} \\
\hline

\multirow{4}{*}{4 humans}    & SR (\%) $\uparrow$ &  92 & 70 & \textbf{100} \\
        & Dist. (m) $\downarrow$ & 0.41 & 0.79 & \textbf{0.07} \\
        & Cont. $\uparrow$ & 0.94 & 0.95 & \textbf{0.96} \\
        & Pene-H. $\downarrow$ & 60.4 & 41.7 & \textbf{0} \\
\hline

\multirow{2}{*}{Diversity} & SR (\%) $\uparrow$ & 96 & 84 & \textbf{97}\\
            & Std Dev $\uparrow$ & 0.987 & 1.05 & \textbf{1.21}\\
\hline
 \end{tabular}
 \end{adjustbox}

 \label{tab:eval}
 \vspace{-3mm}
\end{table}

We assess CAMPs' generalizability in dynamic scenes, including scenes with moving obstacles and scenes with multiple individuals.
In tests with moving obstacles, the obstacle blocks the person's path by moving between the person and the goal. In multiple human test scenes, lines from their starting and goal locations intersect in the middle, requiring solving human-human penetrations. See detail in \cref{sec:s-testscene}.

In Tab.~\ref{tab:eval}, we compare goal-reaching behaviors with two recent baselines: GAMMA~\cite{zhang2022wanderings} and DIMOS~\cite{zhao_synthesizing_2023}.
Baseline methods use navigation meshes and path planning for static scenes, while CAMPs can autonomously avoid dynamic obstacles (Sec.~\ref{sec:3-3}). For fair comparison in dynamic scenes, we extend the baselines by updating navigation meshes and performing on-the-fly path planning at each time step. The tree-based search as in~\cite{zhang2022wanderings} is disabled for all the methods.
\textbf{Metrics}:
(1) \textit{SR}: Success rate for reaching the goal location within a 0.3m threshold.
(2) \textit{Dist.}: Average distance of the final pelvis location to the goal.
(3) \textit{Cont.}: The contact metric~\cite{zhang2022wanderings} that measures foot-floor contact and foot skating.
(4) \textit{Pene-S.}: Percentage of frames with detected human-scene penetration in moving obstacle scenes.
(5) \textit{Pene-H.}: Accurate human-human penetration evaluation metric using COAP~\cite{Mihajlovic:CVPR:2022} in multiple human scenes. 
Please refer to 
\cref{sec:s-metric} for metric details.

CAMPs outperform the two baselines in dynamic scenarios with moving obstacles and multiple humans, exhibiting lower human-scene and human-human penetrations and a higher goal-reaching success rate. 
In multiple human scenarios, we observe that in the baselines, dynamically redoing path planning for each human independently can not effectively solve human-human penetration. In contrast, %
composable CAMPs can generalize well in dynamic settings without using multi-agent RL to synthesize crowd motions.

We assess walking path diversity using the standard deviation of pelvis locations for the same start-target pairs in scenes with a single static box obstacle.
As shown in Tab.~\ref{tab:eval} (Diversity), our approach does not require a pre-computed global path and allows agents to self-explore without being constrained by predefined paths, achieving higher walking path diversity and success rate. This fosters diverse synthetic data generation via more diverse synthesized motion.

\subsection{Evaluation of Egocentric Sensing}
\label{sec:5-2}
\begin{table}[t]
\caption{Evaluation of egocentric sensing. (Sec.~\ref{sec:5-2})}
\centering
\begin{adjustbox}{width=0.45\textwidth}
 \begin{tabular}{||c c c||} 
 \hline
  Method (sensing range) & SR (\%) $\uparrow$ &   Dist. (m) $\downarrow$  \\ [0.5ex] 
 \hline\hline
  Local map~\cite{zhao_synthesizing_2023} (0.8 m) &  78 &  0.35 \\ 
  Local map* (7 m)  &  4  &  3.04\\
 Egocentric sensing (ours) (7 m)   &  \textbf{95} & \textbf{0.12}  \\
 \hline
 \end{tabular}
 \end{adjustbox}
 \label{tab:egosensing}
 \vspace{-3mm}
\end{table}

We assess the exploration ability of our egocentric sensing $\mathcal{E}_t$ in Replica~\cite{replica19arxiv} scenes. In Tab.~\ref{tab:egosensing}, we replace $\mathcal{E}_t$ with a local map~\cite{zhao_synthesizing_2023} in our state $\mathbf{s}_t$, following their encoding method. Relying on local information can trap agents in local optima, e.g., walls beyond their sensing range, resulting in lower SR. Our egocentric sensing acts as a depth proxy, allowing the agent to avoid local optima, explore more effectively than local maps~\cite{zhao_synthesizing_2023} or scandots~\cite{DBLP:conf/corl/AgarwalKMP22}, and achieve higher SR. In addition, our compact representation is more scalable as the sensing range increases, while quadratic local map growth can hinder the policy network's learning.

\subsection{Ablation Studies}
\label{sec:5-3}
\begin{table}[t]
\caption{Ablation studies. \textit{Note: in our observation, $\|$VP$\|_2 > 15$ indicates abnormal human poses.} (Sec.~\ref{sec:5-3})}
\centering
\begin{adjustbox}{width=0.45\textwidth}
 \begin{tabular}{||c c c c||} 
 \hline
   & SR (\%) $\uparrow$ &   $\|$VP$\|_2$ $\downarrow$ & $\cos (\vv{\mathbf{v}}, \vv{\mathbf{a}})$ $\uparrow$ \\ [0.5ex] 
 \hline\hline
  Egocentric depth & 8 & 13.64 & 0.049 \\ 
  No pretraining & 90 & 28.77 & 0.918\\
   No attention reward &  90 &  12.26 & 0.891 \\
  Our policy & \textbf{92} & \textbf{10.57} & \textbf{0.940}\\
 \hline
 \end{tabular}
 \end{adjustbox}
 \label{tab:ablation}
 \vspace{-3mm}
\end{table}

We compare our policy with several ablations in Tab.~\ref{tab:ablation}: \newline \textbf{Egocentric depth}: an ablation training an egocentric depth image-based policy without the depth sensing proxy. Egocentric depth images are encoded with a CNN; \newline 
\textbf{No pretraining}: an ablation training collision avoidance in crowded scenes with strict penetration termination directly; \newline 
\textbf{No attention reward}: an ablation for the viewing direction. 

We assess pose naturalness with the maximum pose embedding norm encoded with VPoser~\cite{SMPL-X:2019} and evaluate the attention reward with the cosine similarity between the viewing direction $\protect\vv{\protect\mathbf{v}}$ and the attention direction $\protect\vv{\protect\mathbf{a}}$ (Sec.~\ref{att_rew}).

Directly training RL with egocentric depth images is ineffective due to our high-dimensional action space, emphasizing the value of the compact egocentric sensing representation. Training agents with strict penetration constraints in crowded scenes directly can result in exploring unreasonable action subspaces, given its unbounded Gaussian nature, leading to unrealistic human poses, highlighting the effectiveness of our two-stage RL training scheme. Without the attention reward, the virtual human's capability to attend to a specific direction decreases. All ablation studies are evaluated in Replica.
See visuals in Supp. Vid. and 
\cref{sec:s-ablate}.

\subsection{Mapping, Localization, and Tracking for HMD}
\noindent \textbf{Mapping and localization.}
\label{lamar}
LaMAR~\cite{sarlin2022lamar} is the first mapping and localization benchmark dataset for AR in large-scale scenes.  
Despite over a year of extensive data collection, the dataset still lacks exhaustive scene coverage, especially in large open spaces.
\methodname~can let virtual humans explore large-scale scenes, render dense egocentric views, and build a more complete SfM map by %
extracting image feature points with SuperPoint~\cite{detone2018superpoint} and matching images with SuperGlue~\cite{sarlin2020superglue}. 
Despite synthetic images being noisier due to scene quality, SuperGlue~\cite{sarlin2020superglue} matching can filter out noisy feature points and yield reliable matches.

In Tab.~\ref{tab:lamar}, we evaluate \methodname~by assessing the localization recall at ($1^{\circ}, 10cm$) on the validation set in a lobby of $\sim$120 sqm of the LaMAR CAB location. In addition, we report the number of triangulated 3D points (\#P3D) and track length. 
\methodname~improves the 3D reconstruction by yielding more points for a slightly improved track length and also a significantly better localization performance compared to using the real data only. Ng et al.~\cite{ng2021reassessing} augments mapping images by perturbing real-world camera poses with noise, which may generate unrealistic camera poses (e.g., stuck in a wall or facing the ceiling), limiting egocentric localization effectiveness. Their method also assumes the availability of initial camera poses, which may not always be feasible.
In contrast, \methodname~augments by virtual humans \textit{randomly} exploring scenes. Our approach holds promise for creating AR mapping and localization datasets for digital twin scenes without manual data collection, providing enhanced privacy preservation, e.g. no need for anonymization.
Refer to \cref{sec:s-lamar} for visualization and implementation details.

\begin{table}[t]
\caption{Mapping and localization evaluation. We augment LaMAR with the same amount of images (248 frames) and report the localization recall at ($1^{\circ}, 10cm$) on the validation set. \methodname~achieves the highest track length and recall. (Sec.~\ref{lamar})}
\centering
\begin{adjustbox}{width=0.45\textwidth}
 \begin{tabular}{||c c c c||} 
 \hline
   & \#P3D $\uparrow$ & Track length $\uparrow$ &  Recall (\%) $\uparrow$ \\ [0.5ex] 
 \hline\hline
 LaMAR   & 1929739 & 5.1946 & 66.9\\ 
 Ng et al.~\cite{ng2021reassessing}  & \textbf{1937758} & 5.1940 & 74.9 \\
 EgoGen & 1936169 & \textbf{5.2105} & \textbf{76.7} \\
 \hline
 \end{tabular}
 \end{adjustbox}
 \label{tab:lamar}
 \vspace{-2mm}
\end{table}

\begin{table}[t]
\caption{Egocentric camera tracking evaluation of models trained with and without synthetic data from \methodname. (Sec.~\ref{EgoEgo})}
\centering
\begin{adjustbox}{width=0.45\textwidth}
 \begin{tabular}{||c c c c||} 
 \hline
   & Pose $\downarrow$ & Rotation $ \downarrow$ & Transl $ (mm) \downarrow$  \\ [0.5ex] 
 \hline\hline
 Scratch   & 1.83 & 0.74 & \textbf{1303} \\ 
 + \methodname~pretrain & \textbf{1.67} & \textbf{0.62} & 1305 \\
 \hline
 \end{tabular}
 \end{adjustbox}
 \label{tab:egoego}
 \vspace{-4mm}
\end{table}

\noindent \textbf{Egocentric camera tracking.}
\label{EgoEgo}
Egocentric camera tracking for HMD aims to yield device pose trajectories in 3D scenes given egocentric video observations.
Recovering camera poses from monocular RGB videos using SLAM \cite{teed2021droid} is a challenging and ill-posed problem due to scale ambiguity. EgoEgo \cite{egoego} leverages the knowledge of human motion to address the egocentric HMD tracking problem. Specifically, EgoEgo trains a 
neural network to infer the translation scaling and rotations from egocentric videos, which improves the HMD tracking performance. 
However, training this model requires jointly captured data of ground truth HMD trajectories and egocentric videos, which are costly to collect.
We address this limitation by using \methodname~to synthesize quantities of egocentric videos with accurate camera trajectories to pretrain the model, which proves to improve the tracking performance on real data. 
We conduct experiments on the GIMO \cite{zheng2022gimo} dataset that contains $\sim$200 short sequences of paired motion-video data in 19 scenes. 
Using~\methodname, we synthesize $\sim$4k sequences of human movements in their scenes and render corresponding egocentric videos using the same camera intrinsic as GIMO and the embodied camera placement described in Sec.~\ref{sec:cam_place}.
We also slightly perturb the camera placement location and orientation to simulate the diversity of how people wear HMDs in real data and avoid overfitting to one specific camera placement.
We first pretrain the model with synthetic data generated by \methodname, then finetune it on the real GIMO data.
Tab.~\ref{tab:egoego} shows the egocentric camera tracking performances for models trained with and without synthetic data. 
Definitions of evaluation metrics can be found in \cref{sec:s-egoego}.
The finetuned model benefits from \methodname~synthetic data and predicts more accurate camera poses compared to the model trained using real data only.

\subsection{Human Mesh Recovery from Egocentric views}
\label{exp-egobody}

Human mesh recovery (HMR) is the key to human behavior understanding from the egocentric view, thus crucial for applications in robotics and AR/VR. Given an egocentric RGB or depth image of a target subject, HMR aims to reconstruct the subject's 3D body pose and shape. 
However, acquiring and annotating real-world data is expensive, demanding, and time-consuming, with egocentric data being particularly scarce.
EgoBody~\cite{Zhang:ECCV:2022} is a recent egocentric dataset capturing two-people interactions, with egocentric depth/RGB frames annotated with SMPL-X body meshes. EgoBody provides $\sim$180k egocentric RGB frames, and merely $\sim$23k depth frames due to the low frame rate of the depth sensor, with $\sim$90k/$\sim$10k in the RGB/depth training set. Such limited data is insufficient to train a learning-based model from scratch.
In contrast, with \methodname, 
large-scale synthetic egocentric data can be generated in a time-efficient way. 
We leverage \methodname~to generate quantities of training frames (300k RGB, 105k depth) of humans moving in EgoBody 3D scenes, rendered from the egocentric view, and annotated by SMPL-X parameters of the target subject. Specifically, RGB images are rendered with lifelike human body textures and 3D clothing, with random lighting. 

With the recent HMR regressor, ProHMR~\cite{kolotouros2021probabilistic}, %
we show that pre-training with our synthetic data from \methodname~enhances the existing method's capability to generalize on real-world scenarios.
Evaluated on the real-world EgoBody test set, we compare two training schemes: (1) trained from scratch on the real-world EgoBody training set (``-scratch''), and (2) pre-trained on synthetic data from \methodname~and fine-tuned on the real-world EgoBody training set (``-ft'').

\textbf{HMR from depth.}
As no existing methods were proposed for depth-based HMR task, we adapt ProHMR~\cite{kolotouros2021probabilistic} to the depth input by changing the channel number of the first convolution layer.
To mimic real-world sensor noise, synthetic noise~\cite{handa:etal:2014} is added to the rendered depth.
G-MPJPE is additionally reported for depth-based HMR as depth images provide global information.
As shown in Tab.~\ref{tab:hmr}, compared to the model trained only with a limited amount of real-world data (Depth-scratch), errors are significantly reduced for the model pre-trained with our large-scale synthetic data (Depth-ft), in terms of global translation (22.9\% lower G-MPJPE), local pose (20.7\% lower MPJPE), and body shape (19.5\% lower V2V). %

\textbf{HMR from RGB.} 
For training with RGB images, we apply various data augmentation techniques similar to~\cite{Black:CVPR:2023}. 
Tab.~\ref{tab:hmr} indicates that the RGB-based model pre-trained with large-scale synthetic data (``RGB-ft'') also outperforms the model trained only on real-world data (``RGB-scratch''), for both body pose and shape accuracy.

The enhanced performance highlights that \methodname's synthetic data effectively compensates for the lack of real-world training data, boosting the performance of current methods when test on real-world data. We will release both of our synthetic EgoBody datasets. 
See \cref{sec:shmr} for dataset statistics, qualitative visualizations, and training details.

\begin{table}[t]
 \caption{Evaluation of HMR on EgoBody test set. ``*-scratch'' denotes the model trained from scratch with the Egobody training set, and ``*-ft'' denotes the model pre-trained with \methodname~synthetic data. The units for all metrics are in \textit{mm}. (Sec.~\ref{exp-egobody}) \vspace{-2mm}}
\centering
\begin{adjustbox}{width=0.45\textwidth}
 \begin{tabular}{||c c c c c||} 
 \hline
   & G-MPJPE $\downarrow$ & MPJPE $\downarrow$ & PA-MPJPE $\downarrow$ & V2V $\downarrow$ \\ 
 \hline\hline
 Depth-scratch   & 117.7 & 82.2 & 54.1 & 100.6\\ 
 Depth-ft  &\textbf{90.7}   &\textbf{65.2}  &\textbf{47.3}  &\textbf{81.0}  \\
 \hline\hline
 RGB-scratch & - & 90.7 & 59.9 & 102.1\\
 RGB-ft & - & \textbf{85.3} & \textbf{56.2} & \textbf{97.2} \\
 \hline
 \end{tabular}
\end{adjustbox}
 \label{tab:hmr}
 \vspace{-4mm}
\end{table}

\section{Conclusion and Future Work}

We propose a novel egocentric synthetic data generation approach, \methodname, that uses embodied sensors, a parametric body model, and a generative egocentric perception-driven human motion synthesis method to create egocentric training data with accurate and rich ground truth annotations. 
By integrating deep reinforcement learning and collision-avoiding motion primitives with egocentric depth proxy, \methodname~synthesizes robust human motion and emergent multi-agent behaviors. This paves the way to an efficient and scalable data generation solution that may have a profound impact on egocentric perception tasks. 

Human-scene interaction in \methodname~is currently coarse. We aim to extend the current method to simulate more detailed human motion driven by egocentric perception, such as hand manipulation, sitting, lying, etc, to facilitate more realistic egocentric synthetic data.
We use fixed attention goals to model human attention.
Predicting human intention through historical egocentric perception and synthesizing viewing directions based on predicted intention holds significant potential.
Synthesizing gaze direction for predicting human intent is valuable but presently hampered by data requirements; we will revisit this when resources allow.

We will explore many other egocentric vision tasks with \methodname~as this area grows rapidly such as social understanding and forecasting.
\methodname~could benefit human-robot interaction, e.g., our generative human motion model and lifelike human appearances can be integrated into \cite{DBLP:journals/corr/abs-2310-13724} to close the sim2real gap for robotic agents further.

\noindent\textbf{Acknowledgements.} This work is supported by the SNSF project grant 200021 204840 and the SDSC PhD fellowship. We sincerely thank Linfei Pan, Yunke Ao, Qianli Ma, and Maxime Raafat for the valuable discussions.
\clearpage
{
    \small
    \bibliographystyle{ieeenat_fullname}
    \bibliography{main}
}

\clearpage
\setcounter{page}{1}
\setcounter{table}{0}
\setcounter{figure}{0}
\setcounter{section}{0}
\maketitlesupplementary

\makeatletter
\renewcommand \thesection{S\@arabic\c@section}
\renewcommand\thetable{S\@arabic\c@table}
\renewcommand \thefigure{S\@arabic\c@figure}
\makeatother

\section{Related Work}
\label{sec:srelated_work}

\methodname~addresses the gap in egocentric synthetic data generation specifically tailored for head-mounted devices, situated at the intersection of three key areas: 1) General synthetic data generation; 2) Egocentric simulation for embodied agents; 3) Human-related synthetic data generation. %
For a more detailed understanding of the distinctions between \methodname~and existing methods, refer to Tab.~\ref{tab:compare}, where these three areas are clearly outlined within distinct blocks. 

In particular, VirtualHome~\cite{puig2018virtualhome, puig2021watchandhelp} also provides rendered egocentric views from head-mounted cameras. However, their egocentric videos lack fluctuating patterns due to the absence of natural human motion; instead, they display robotic-like patterns as Habitat 2.0~\cite{szot2021habitat}. We advance closer to synthesizing more realistic data for head-mounted devices. In addition, they lack a generative human motion model, whereas ours can generate more diverse human motion and trajectories. A very recent work Habitat 3.0~\cite{DBLP:journals/corr/abs-2310-13724} introduced virtual humans to robotic simulation. However, their human locomotion is synthesized by cyclically replaying a walking motion clip from MoCap data along a pre-calculated path with rigid rotations to transition to the next walking direction. %
Both VirtualHome and Habitat 3.0 have a limited number of human agents and fall short in representing diverse human characters, with limitations in body shapes, ethnic variation, and clothing options. UnrealEgo \cite{hakada2022unrealego} is a large-scale naturalistic dataset for egocentric 3D human motion capture, which employs downward-facing cameras and replies on mocap data. It is used to estimate the human body pose of the camera wearer. In contrast, we employ front-facing cameras and perform generative human motion synthesis for autonomous virtual humans.

Our synthetic data generation achieves increased diversity by incorporating a walking path-free generative human motion model, diverse body shapes, various body textures, and varied 3D textured clothing.

\begin{table*}[htbp]
\caption{Comparison of existing synthetic datasets or generators. (Sec. \ref{sec:srelated_work})}
\centering
\begin{adjustbox}{width=0.95\textwidth}
    \rowcolors{2}{white}{gray!20}
 \begin{tabular}{||c c c c c c c c||} 
 \hline
    & Domain & Egocentric & Head-mounted & \makecell{Multi-Camera \\ rigs} & Virtual Humans &  \makecell{Automated \\ Clothing Simulation}  & \makecell{Generative Human \\ Motion Synthesis} \\
 \hline\hline
  Kubric~\cite{greff2021kubric} & Scattered Objects & \xmark &  \xmark & \xmark & \cmark &  \xmark & \xmark  \\
  PointOdyssey~\cite{zheng2023point} & Point Tracking &\cmark &  \cmark & \xmark & \cmark & \xmark  & \xmark  \\
  InfiniGen~\cite{infinigen2023infinite} & Natural &  \xmark &  \xmark &  \xmark &  \xmark  &  \xmark  &  \xmark   \\
  RoboGen~\cite{wang2023robogen}  & Robotics & \xmark & \xmark &  \xmark &\xmark  & \xmark  &  \xmark \\
  UniSim~\cite{yang2023learning} & Real-world Interaction & \cmark & \cmark & \xmark & \xmark & \xmark  &  \xmark\\
  UnityPerception~\cite{unity-perception2022,ebadi2022psphdri} & \makecell{Object Detection\\Pose Estimation}  & \xmark & \xmark & \xmark & \cmark & rigged clothing & \xmark \\
 uHumans2~\cite{Rosinol20rss-dynamicSceneGraphs} & Scene Graphs & \cmark & \xmark &\xmark & \cmark  & \xmark  & \xmark \\
  \hline \hline
  Carla~\cite{dosovitskiy2017carla} & Driving & \cmark & \xmark & \xmark& \cmark & rigged clothing & \xmark \\
  VirtualHome~\cite{puig2018virtualhome, puig2021watchandhelp}  & Household Simulation & \cmark & \cmark & \xmark &  \cmark & rigged clothing & \xmark \\
  VRKitchen~\cite{VRKitchen}  & Cooking Simulation & \cmark & \cmark & \xmark &  \xmark & \xmark  & \xmark \\
  Habitat 2.0~\cite{szot2021habitat}& Embodied Robots &  \cmark  &   \xmark & \xmark & \xmark & \xmark  &  \xmark \\
  Habitat 3.0~\cite{DBLP:journals/corr/abs-2310-13724} &\makecell{Human-robot\\ Interaction} &  \cmark  & \xmark &  \xmark  & \cmark & \xmark  &  \xmark \\
  \hline \hline
  UnrealEgo~\cite{hakada2022unrealego} & Ego-pose Estimation &  \cmark  & \cmark & \cmark & \cmark & rigged clothing &  \xmark \\
  GTA-Human~\cite{cai2021playing}   & Pose Estimation & \xmark & \xmark & \xmark  & \cmark & \xmark & \xmark \\
    BEDLAM~\cite{Black:CVPR:2023} &  Pose Estimation &   \xmark & \xmark & \xmark & \cmark & \xmark  & \xmark  \\
  SynBody~\cite{yang2023synbody} & Pose Estimation & \xmark & \xmark & \xmark & \cmark & \xmark  &  \xmark \\
  ADT~\cite{DBLP:journals/corr/abs-2306-06362} & Digital Twin &  \cmark & \cmark & \cmark & \xmark & \xmark  &  \xmark \\
  \hline \hline
  EgoGen (ours) & \makecell{Head-mounted\\Devices} & \cmark & \cmark & \cmark & \cmark & \cmark  &  \cmark \\
 \hline
 \end{tabular}
 \end{adjustbox}
 \label{tab:compare}
\end{table*}
\section{Ego-Sensing Driven Motion Synthesis}

\subsection{Egocentric Sensing Calculation}
\label{sec:s2-1}
As a compact and cheap-to-compute representation of depth maps, egocentric sensing resembles the calculation of depth information but is simplified into 2D.

As shown in~\cref{fig:joints}, the location of the egocentric camera is the midpoint of two eyeballs, and the viewing direction $\protect\vv{\protect\mathbf{v}}$ is visualized as the red arrow. 
$N$ rays are cast from the location of the egocentric camera, with the central direction of these rays determined by the 2D projection of $\protect\vv{\protect\mathbf{v}}$.
The starting points of these rays are identical, while their endpoints form a semicircle in front of the virtual human, representing the field of view $[\theta_{min}, \theta_{max}]$ to the human. Each ray has the potential to extend infinitely.
In our implementation, $N=32$, $\theta_{min}=-90^{\circ}$, $\theta_{max}=90^{\circ}$.

\begin{figure}[htbp]
    \centering
    \setlength{\tabcolsep}{0.0130\linewidth}
    \includegraphics[width=1.0\linewidth]{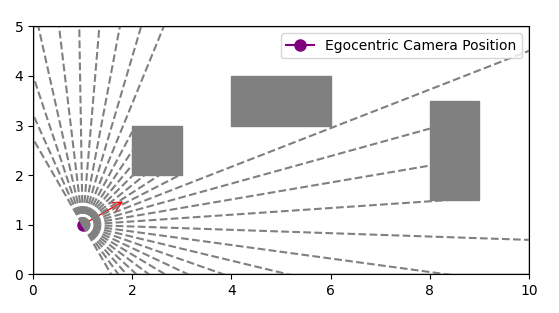}%
    \caption{The 2D projection of the egocentric camera location is represented by the purple point, while the 2D projection of the viewing direction $\protect\vv{\protect\mathbf{v}}$ is indicated by the red arrow. The field of view changes due to the head pose.
      \label{fig:egosensing}}
\end{figure}

The 2D collision detection of rays leverages the 2D layout of the 3D scene. For illustration purposes, we simplify the obstacles in 3D scenes with grey rectangles and visualize the collision detection of rays in~\cref{fig:egosensing}. The egocentric sensing encodes the simplified depth of obstacles.

\subsection{Reward, Weighting, and Training Detail}
\label{sec:s-reward}
In our motion primitive environment, we design an intuitive set of rewards to encourage the agent to perform realistic human motions (all vectors are normalized in the following equations):

    \textbf{Attention reward} that encourages the human to look at the goal:
\begin{equation}
r_{attention}=\frac{\langle \protect\vv{\protect\mathbf{v}}, g - h \rangle + 1}{2},
\label{equ:att_reward}
\end{equation}
where $ \protect\vv{\protect\mathbf{v}}$ denotes the viewing direction, $g$ and $h$ denote the goal location and current head location and $\langle \cdot \rangle$ denotes the inner product.

    \textbf{Foot contact reward} $r_{cont.}$ contains two components: Foot floor distance reward and foot skating reward.
    \begin{equation}
    r_{cont.} = r_{floor} + r_{skate}
    \end{equation}
    \begin{equation}
    r_{floor} = e^{-(|\min_{x \in F} x_z| - 0.02)_+},
    \end{equation}
    \begin{equation}
    r_{skate} =  e^{-(\min_{x \in F}\|x_{vel}\|_2 - 0.075)_+},
    \end{equation}

where $F$ denotes foot markers, $x_z$ denotes the marker height, $x_{vel}$ denotes the marker velocity, and $(\cdot)_+$ denotes clipping negative values. There are tolerance thresholds of 0.02m for foot-floor distance and 0.075m/s for skating.

    \textbf{Goal Distance reward} that encourages the agent to get closer to the goal at each step:
\begin{equation}
r_{dist} = d^{t-1} - d^t,
\label{equ:dist_reward}
\end{equation}

Here $d^t$ denotes the body-goal distance at step $t$. 

    \textbf{Body orientation reward} that encourages the body forward direction to be aligned with the goal location direction:
     \begin{equation}
r_{ori}=\frac{\langle o_b, g - p \rangle + 1}{2},
\label{equ:ori_reward}
\end{equation}
where $o_b$ denotes the body forward orientation, $g$ and $p$ denote the goal location and current pelvis location, and $\langle \cdot \rangle$ denotes inner product. Different from the {\it attention reward} that drives the head motion, this penalizes backward movement toward the goal.

    \textbf{Penetration reward} that penalizes the intersection of the human body and obstacles. We use different penetration rewards in different settings.

    When training in sparse scenes, e.g., a single static box obstacle, penetration detection is simplified into 2D to accelerate calculation:
    \begin{equation}
r_{pene}^{sparse} = 
    \begin{cases}
        0.05, & |\mathcal{M}_0 \cap \mathcal{B}_{xy}(X)| < thres\\
        0, & otherwise
    \end{cases}
\end{equation}

where $\mathcal{M}_0$ denotes the non-walkable cells on the ground plane, $\mathcal{B}_{xy}(.)$ denotes the 2D bounding box of the body markers $X$, $\cap$ denotes their intersection, and $|\cdot|$ denotes the number of non-walkable cells within the bounding box of the human. $thres$ is set to 3 and the cell dimension is $0.1m\times0.1m$.

When training in crowded scenes, we use the signed distance field ($\Psi_O$) for precise penetration detection:
    \begin{equation}
r_{pene}^{crowded} = e^{-\frac{1}{T}\sum_{t=1}^{T}\sum_{i=1}^{|V|}|(\Psi_O(\mathbf{v}_{ti}))_{-}|}\\
\end{equation}

where $|V|$ denotes the number of SMPL-X mesh vertices, $T$ denotes the number of frames in our motion primitive ($T=20$), $\mathbf{v}$ denotes SMPL-X mesh vertex, and $(\cdot)_{-}$ denotes clipping positive values. The penetration reward penalizes body vertices with negative SDF values within a motion primitive.

    \textbf{Pose reward} that penalizes generating unrealistic human poses using VPoser \cite{SMPL-X:2019} body pose prior:  
\begin{equation}
    r_{pose} =  \begin{cases}
        0.05, & \|VP\|_2 < thres\\
        0, & otherwise
    \end{cases}
    \end{equation}
where $\|$VP$\|_2$ denotes the pose embedding inferred by the VPoser encoder $\mu(\cdot)$, where $\|$VP$\|_2 = |\mu(\theta)|_2$.
$\theta$ denotes the SMPL-X body pose parameter representation. The VPoser pose prior learns a probabilistic pose distribution where vectors closing to zero have a high probability and correspond to realistic human poses. In our observation, $\|$VP$\|_2 > 15$ produces unrealistic human poses. $thres$ is set to 11. 

    \textbf{Success reward} for reaching the goal location:
    \begin{equation}
r_{succ} = 
    \begin{cases}
        1, & d < thres\\
        0, & otherwise
    \end{cases}
\end{equation}
where $d$ is the body-goal distance. $thres$ is set to 0.1.

The weights for each reward are listed in~\cref{tab:1}. The weighting of each reward is determined according to the reward value. For example, the goal distance reward measures the distance change in one motion primitive spanning 0.5s, which is approximately 10 times smaller than other rewards. As a result, its weight is 10 times bigger than others. We observe high foot skating weight helps to reduce foot skating. Higher success rewards encourage the agent to reach the goal. But on the other hand, the weight can not be too big. Because we did not do reward normalization, too large values may lead to big errors in value estimation and training instabilities.

\begin{table}[htbp]
\centering
 \begin{tabular}{||c c||} 
 \hline
  Reward & Weight \\ [0.5ex] 
 \hline\hline
  Foot floor distance  & 0.1 \\ 
  Foot skating & 0.3 \\
  Goal distance & 1 \\
  Body orientation & 0.1 \\
  Attention & 0.3 \\
  Penetration pretraining & 1 \\
  Penetration finetuning & 0.1 \\
  Pose & 0.1 \\
  
  Success & 0.5 \\
 \hline
 \end{tabular}
 \caption{Reward weights.}
 \label{tab:1}
\end{table}

\textbf{Penetration Termination.} We terminate an episode due to penetration using different criteria. In sparse scenes, an episode is terminated if $r_{pene}^{sparse} = 0$.

As mentioned in \cref{sec:3-2} in the main paper in crowded scenes, we employ a two-stage RL training scheme. In stage \RNum{1}, we pretrain the policy with a penetration weight of $w_{r_{pene}}=1$ to more effectively encourage the virtual human to avoid obstacles and explicitly \textbf{not} perform penetration termination.
After convergence, in stage \RNum{2}, we proceed to fine-tune the policy with a strict penetration termination using a reduced penetration weight of $w_{r_{pene}}=0.1$. Penetration detection involves considering the maximum number of body vertices in penetration within a motion primitive. An episode is terminated if:
    \begin{equation}
    \max_{t} \sum_{i=1}^{|V|}|(\Psi_O(\mathbf{v}_{ti}))_{-}| \geq thres
\end{equation}
where $thres$ is set to 40.

This design has several reasons: 1) Our action space is an unbounded Gaussian, direct training with strict termination can lead the policy to explore unreasonable spaces and produce unrealistic human poses, see~\cref{fig:weirdpose} for illustration. 2) Reducing penetration weight during fine-tuning can amplify the significance of the goal-reaching weight, encouraging goal-reaching behaviors.

\subsection{PPO}
\label{sec:s-ppo}
Our PPO implementation is based on Tianshou~\cite{tianshou}.
We list the hyperparameters of PPO in~\cref{tab:2}. $c_1, c_2$ are defined in \cref{sec:3-2} in the main paper. ``Repeat per Collect'' is the training iterations with the same collected rollouts.

\begin{table}[htbp]
\centering
 \begin{tabular}{||c c||} 
 \hline
  Param & Value \\ [0.5ex] 
 \hline\hline
  Learning Rate  & 3e-4 \\ 
  $\gamma$ Discount & 0.99 \\
  PPO Clip Threshold & 0.1 \\
  Repeat per Collect & 1\\
  Value Function Coefficient ($c_1$) & 1 \\
  Entropy Coefficient ($c_2$) & 0.01\\
  GAE ($\lambda$) & 0.95 \\
  Max Grad. Norm & 0.1 \\
  
 \hline
 \end{tabular}
 \caption{PPO hyperparameters.}
 \label{tab:2}
\end{table}

The majority of the hyperparameters were set to their default values. To note, adopting smaller values of ``PPO Clip Threshold'' and ``Repeat per Collect'' will not update parameters too drastically and thus stabilize the training.
We use advantage normalization without value function clipping.

Another trick we adopted is that we performed the last policy layer weight scaling, which makes initial actions close to the standard normal distribution, which can boost the performance \cite{DBLP:journals/corr/abs-2006-05990}.

The training time is roughly 20 hours on a GeForce RTX 3090 GPU with batch size 256, 20000 steps per epoch. 

The key difference between our environment with others is that our action space is not strictly bounded. The motion primitive model $\mathcal{P}$ is based on VAE and is pretrained with a KLD loss w.r.t. a standard normal distribution. %
As a result, we do not do any action scaling or clipping during training.
Due to the nature of our action space, the learned policy can deviate too much from the standard normal distribution.
As a result, we select the best model using the best test reward and minimum KL divergence between the learned policy action space and the standard normal distribution.

\subsection{Qualitative Failure Cases}

\begin{figure}[h]
  \centering
  \includegraphics[width=\linewidth]{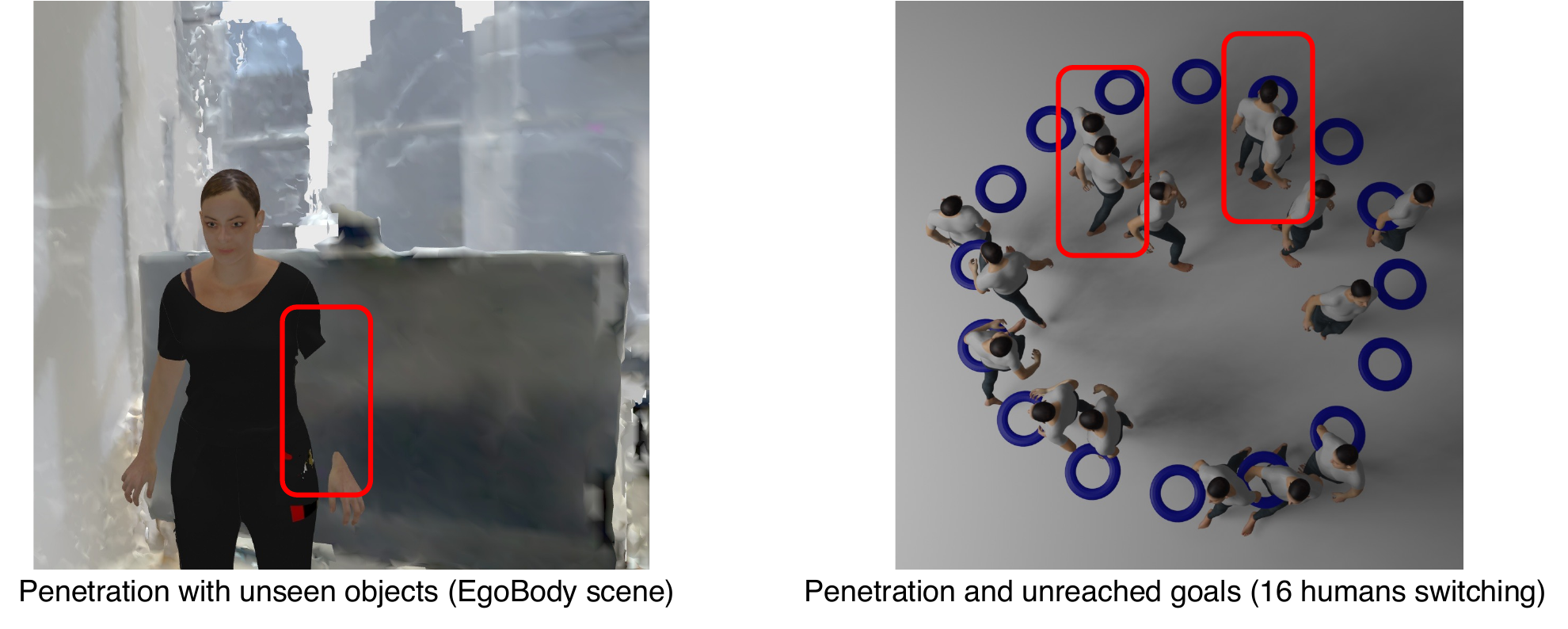}
   \caption{Failure cases.}
   \label{fig:failurecases}
\end{figure}

When evaluation significantly differs from training, failure cases may occur, including penetrations and unreached goals. See \cref{fig:failurecases}.
\section{Egocentric Synthetic Data Generation}
\subsection{Embodied Camera Placement}
\label{sec:supp3-1}

We support various camera placements in \methodname.

\begin{figure}[h]
    \centering
    \includegraphics[width=0.45\linewidth]{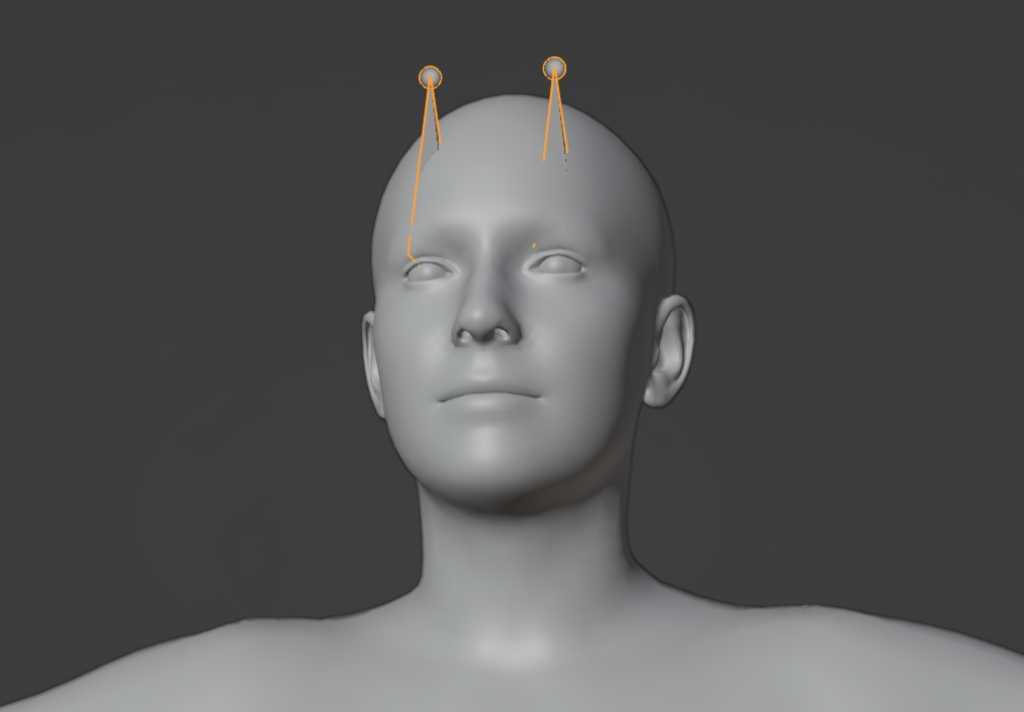}%
    \caption{\label{fig:eyebone}
    Eye bones are located at the eyes and are highlighted with orange edges.
    }
\end{figure}

\begin{figure}[ht]
    \centering
    \setlength{\tabcolsep}{0.0130\linewidth}
    \includegraphics[width=0.47\linewidth]{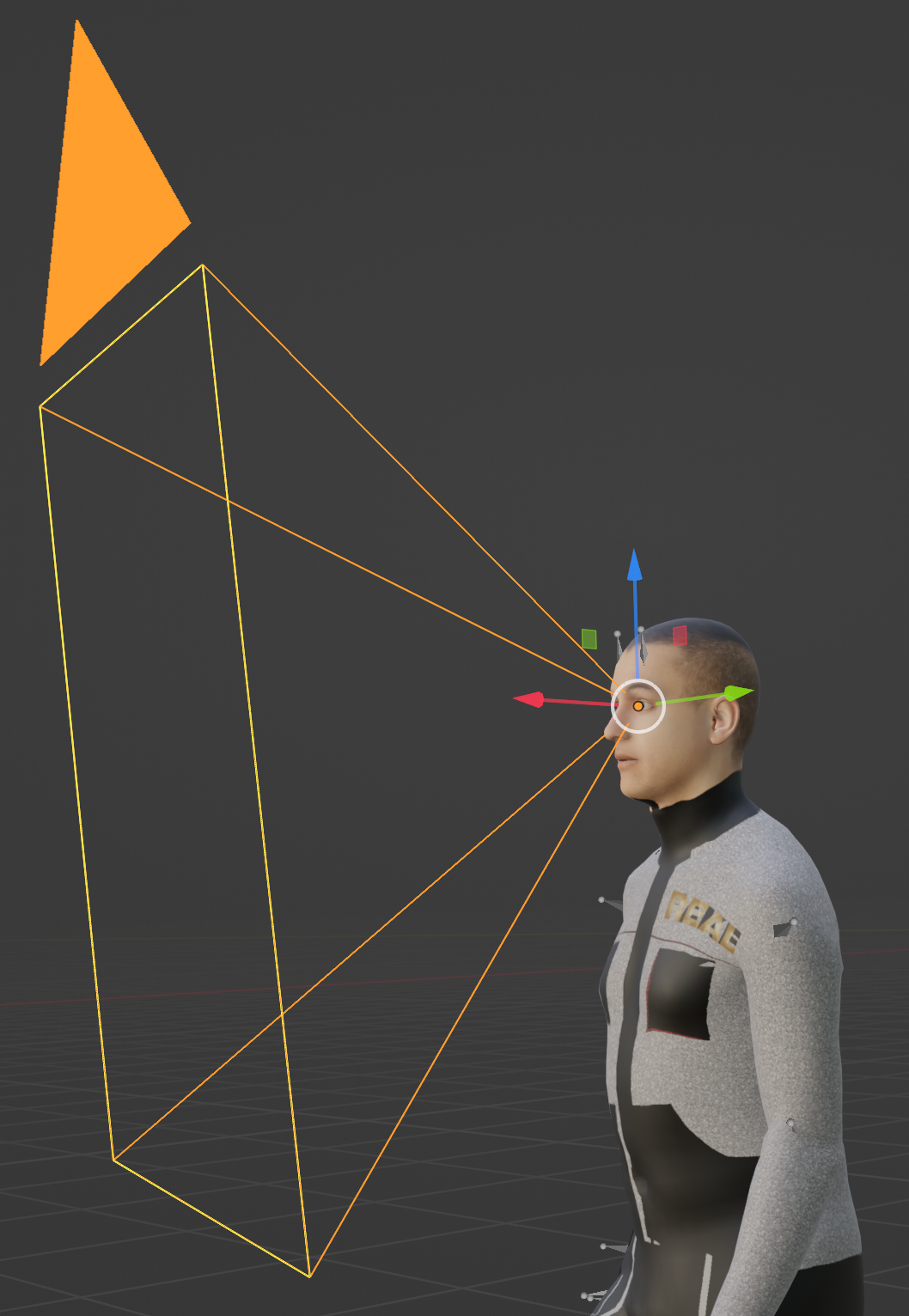}%
    \hspace{1mm}
    \includegraphics[width=0.475\linewidth]{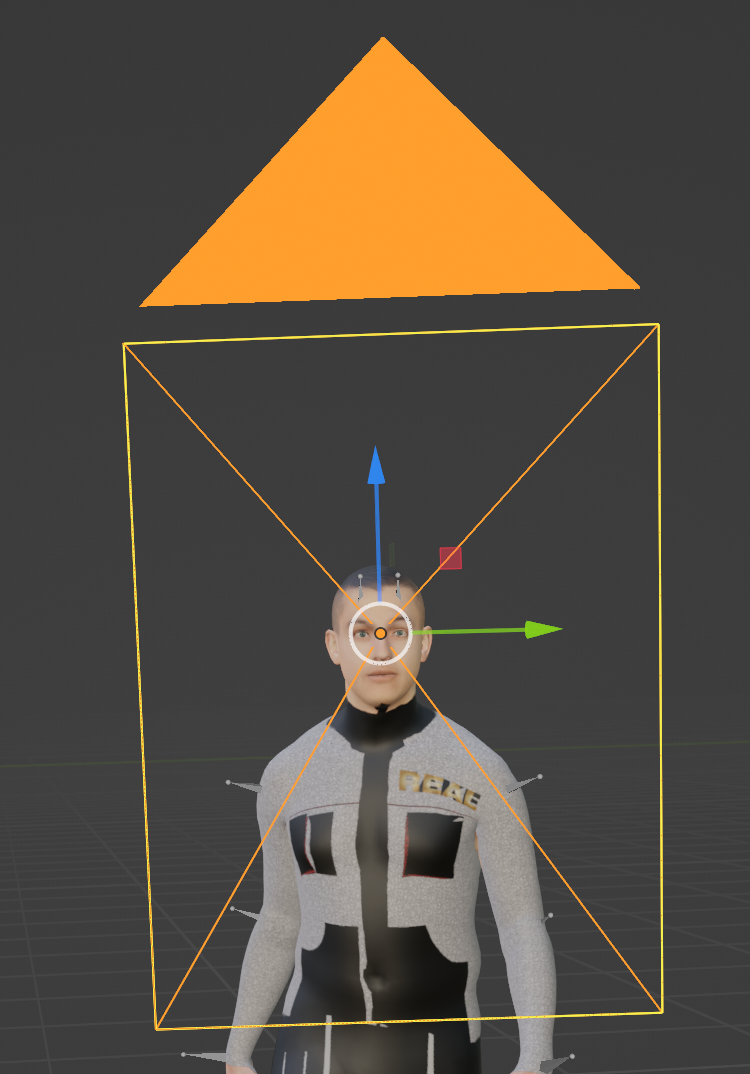}\\
    \caption{\label{fig:joints}
    Illustration of embodied camera placement. The camera axes are determined by 1) The \textcolor{blue}{blue} arrow of the eye bone (from \textit{root} to \textit{tip}); 2) The \textcolor{green}{green} arrow from one eyeball to another; 3) The \textcolor{red}{red} arrow representing the viewing direction $\protect\vv{\protect\mathbf{v}}$. (\cref{sec:s2-1}, \cref{sec:supp3-1})}
\end{figure}

For egocentric sensing-driven motion synthesis (\cref{sec:3} in the main paper), we place one camera at the midpoint of two eyeballs and the viewing direction $\protect\vv{\protect\mathbf{v}}$ is shown in~\cref{fig:joints}. We use the SMPL-X~\cite{SMPL-X:2019} armature in Blender~\cite{bldddd} to calculate $\protect\vv{\protect\mathbf{v}}$. The two eye bones are visualized in~\cref{fig:eyebone}.

\methodname~also supports multi-camera rigs simulation. With the information about the relative poses of cameras within a rig, we have the flexibility to position the camera at various locations on the head. 

To further enhance realism, one may consider various face shapes and physics simulations to place egocentric cameras.
We agree that simulating headset placement and shifting can enhance realism.
However, first, achieving faithful simulation would involve modeling deformable human facial skin and muscles, which is an ongoing research problem.
Second, even with perfect modeling of the camera extrinsic shifting, real-world devices can still have instrumental errors that may offset fine-grained simulation. 
As an alternative, we considered that people can wear the headset with slightly different placements, and introduced noise augmentation to simulate this (See Sec.~\ref{sec:s-egoego}: Egocentric camera tracking).
It might be effective to use aleatoric uncertainty to model headset shifting and instrumental errors together in a Bayesian way.

\subsection{Automated Clothing Simulation}
\label{sec:sclothing-sim}
As shown in~\cref{tab:compare}, many prior works resort to generating synthetic data with rigged clothing, with unrealistic clothing deformations. In contrast, BEDLAM~\cite{Black:CVPR:2023} and Synbody~\cite{yang2023synbody} incorporate physics-based clothing simulation to enhance realism and allow for dressing a diverse range of body shapes in a wide array of clothing. However, their approaches are not scalable for handling arbitrary motion sequences produced by our generative human motion model.

We further automate clothing dynamics simulation with the state-of-the-art clothing simulation network HOOD~\cite{grigorev2022hood}. HOOD treats each garment as a single graph and predicts graph deformations due to both gravity and collisions with the human body mesh.

First, we perform preprocessing on the 3D clothing mesh from~\cite{Black:CVPR:2023} to separate the upper garment and lower garment into distinct clothing meshes because HOOD can not handle disconnected graphs as input.
Second, we sample pose blend shapes, shape blend shapes, and average skinning weights from the closest $n$ SMPL-X mesh vertices in A-Pose, where $n=1$ for tight garments such as pants and $n=1000$ for loose garments such as dresses. We repose the clothing meshes in A-Pose to match the body pose in the first frame of a synthesized motion sequence. Then, for lower garments, the vertices in the top ring are fixed to the body to prevent dropping due to gravity. Finally, we simulate the upper and lower garments separately using HOOD.

\subsection{More Examples of Available Annotations}

\begin{figure}[h]
    \centering
    \setlength{\tabcolsep}{0.0130\linewidth}
    \includegraphics[width=1.0\linewidth]{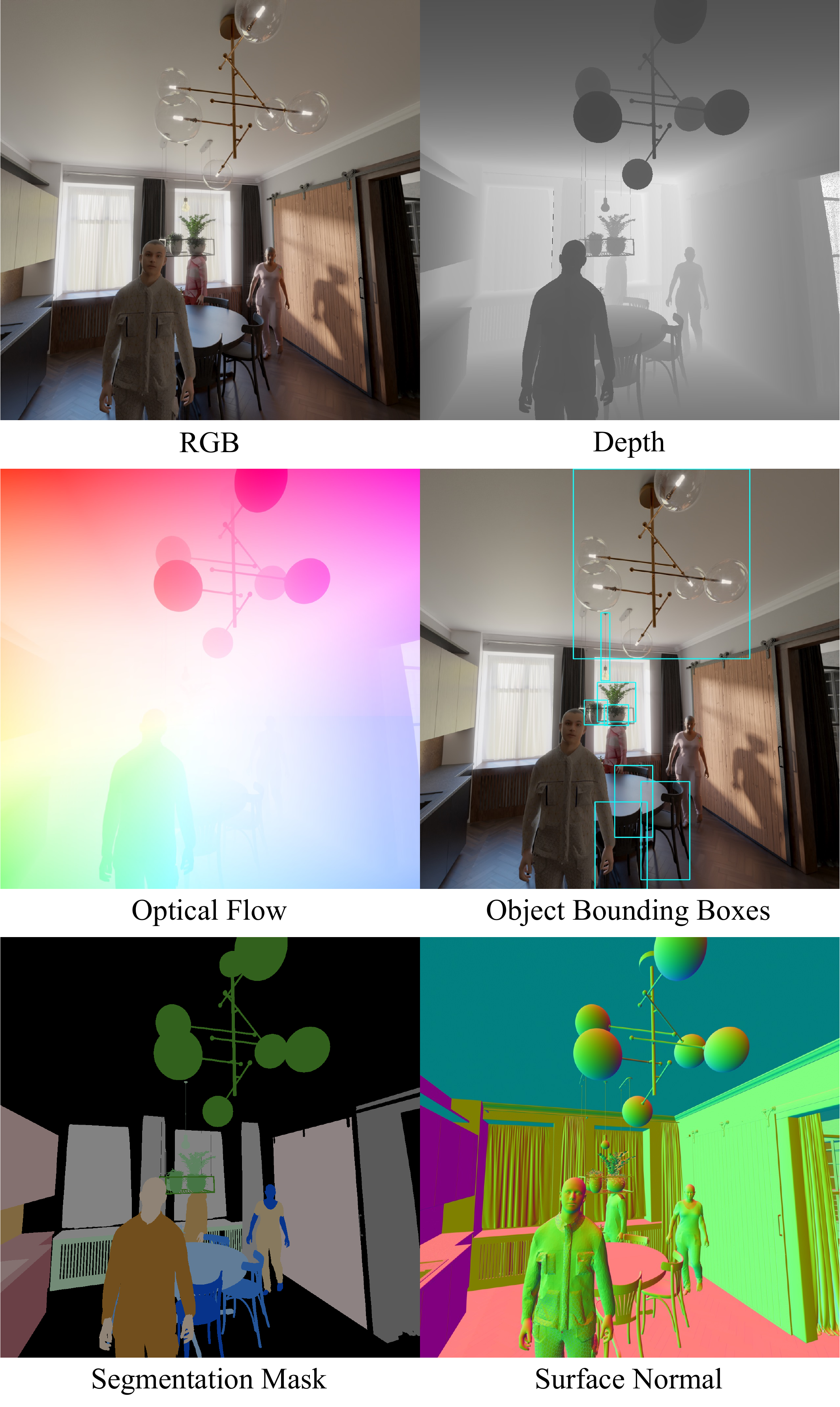}%
    \caption{Ground-truth annotations from perspective cameras.
      \label{fig:annotation}}
\end{figure}

In addition to the fisheye cameras shown in the teaser, here we show more ground-truth annotations with perspective cameras, including RGBD, optical flow, bounding boxes, segmentation masks, and surface normals in~\cref{fig:annotation}.

\section{Experiments}

\subsection{Test Scenarios in Evaluation of CAMPs}

\label{sec:s-testscene}

We provide visualizations of how we built test scenarios. Please refer to Sup. Vid. for qualitative results.

\begin{figure}[htbp]
    \centering
    \setlength{\tabcolsep}{0.0130\linewidth}
    \includegraphics[width=0.9\linewidth]{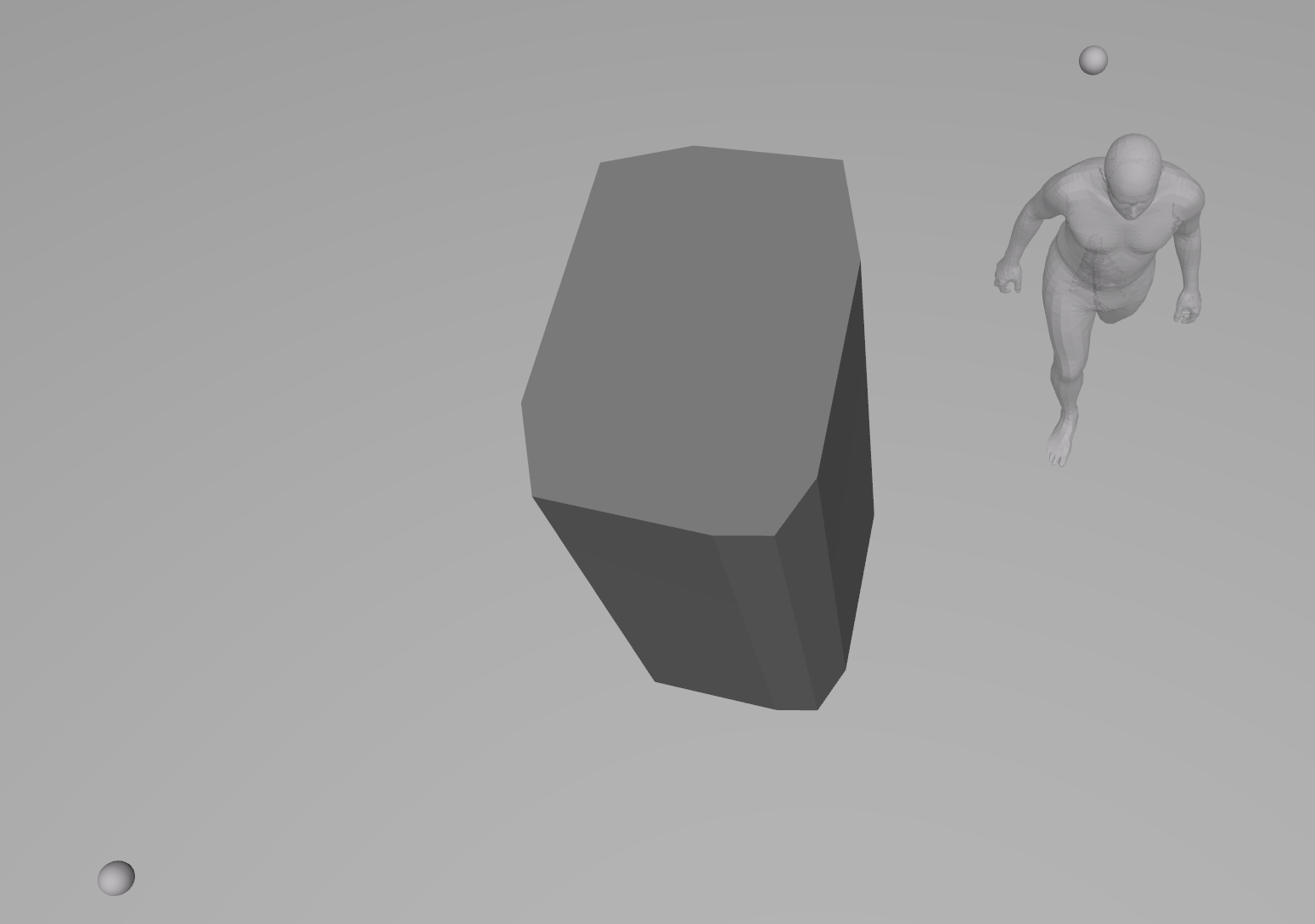}%
    \caption{Moving obstacle test scenario illustration.
      \label{fig:moving}}
\end{figure}

\paragraph{Moving obstacle.} Refer to~\cref{fig:moving} for an illustration of the evaluation in scenes with moving obstacle. %
The moving obstacle will move between the human and its goal location.

\begin{figure}[h]
    \centering
    \setlength{\tabcolsep}{0.0130\linewidth}
    \includegraphics[width=0.9\linewidth]{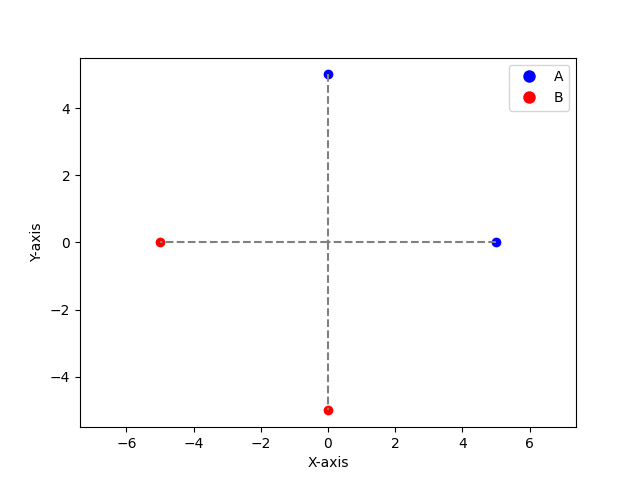}%
    \caption{Multiple humans crowd motion test scenario illustration.
      \label{fig:multiplehuman}}
\end{figure}

\paragraph{Multiple humans.} To visually demonstrate, as depicted in \cref{fig:multiplehuman}, we initiate four virtual humans from distinct points in the figure. We require they walk to the opposite location across the origin, either from A to B or from B to A. There are no other obstacles. Please refer to Sup. Vid. for qualitative results. %

\paragraph{Path diversity.} 
We assess walking path diversity with a static obstacle fixed at the midpoint between the start and target locations. Similar to~\cref{fig:moving}, but the obstacle is not moving anymore.

\subsection{Evaluation Metrics}
\label{sec:s-metric}
The Foot contact metric \cite{zhang2022wanderings} reaches 1 when there is foot-floor contact and no foot skating, defined as: 
\begin{equation}
s_{contact} = e^{-(|\min x_z| - 0.05)_+} \cdot e^{-(\min\|x_{vel}\|_2 - 0.075)_+},
\label{equ:contact}
\end{equation}
where $x_z$ and $x_{vel}$ denote the marker height and velocity, 0.05 and 0.075 are tolerance thresholds, and $(\cdot)_+$ denotes clipping with the lowerbound of 0.

For moving obstacle scenes, we evaluate human-scene penetration by detecting all frames where the floor plane projections of any body parts and obstacles have intersections.
For multiple human scenes, we measure the accurate human-human penetration using the implicit human body occupancy model COAP \cite{Mihajlovic:CVPR:2022}, which predicts the body occupancy given spatial location queries. Since the articulated human bodies are complex and require accurate penetration detection, we detect whether one human collides with other humans by querying its body vertices using the occupancy field of all other humans at that frame and report the occupancy values for human-human penetration.

\subsection{Ablation Studies}
\label{sec:s-ablate}
\begin{figure}[h]
    \centering
    \setlength{\tabcolsep}{0.0130\linewidth}
    \includegraphics[width=0.9\linewidth]{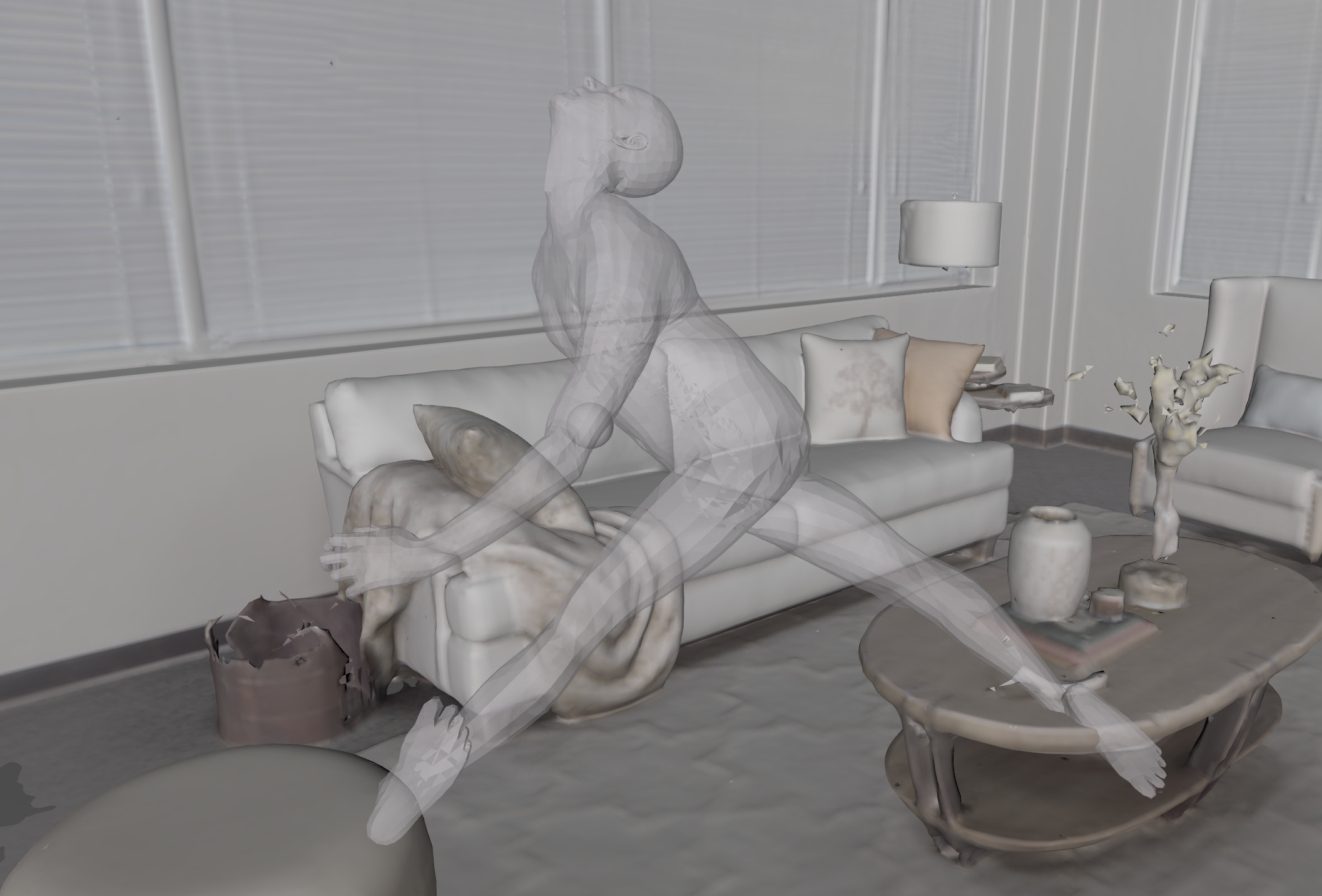}%
    \caption{Ablation of our two-stage RL training. Without pretraining, the model can produce unrealistic human poses. 
      \label{fig:weirdpose}}
\end{figure}

\paragraph{No pretraining.} Without our two-stage RL training scheme, direct penetration termination in crowded scenes will result in unrealistic predicted human poses. As shown in \cref{tab:ablation} in the main paper, where $\|$VP$\|=28.77 > 15$, we provide visualizations of the corresponding unnatural poses in~\cref{fig:weirdpose}. In contrast, our full model works well. Please refer to Sup. Vid.

\section{Egocentric Perception Tasks}
\subsection{Mapping and Localization for AR}
\label{sec:s-lamar}
\begin{figure}[h]
        \centering
        \begin{subfigure}[b]{0.2\textwidth}
            \centering
            \includegraphics[width=\textwidth]{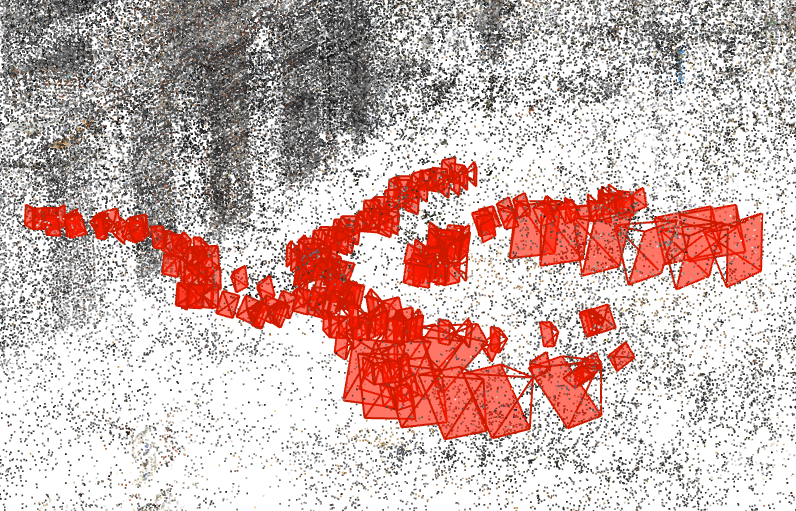}
            \caption{Real-world cameras}
    \label{fig:first}
        \end{subfigure}
        \begin{subfigure}[b]{0.215\textwidth}  
            \centering 
            \includegraphics[width=\textwidth]{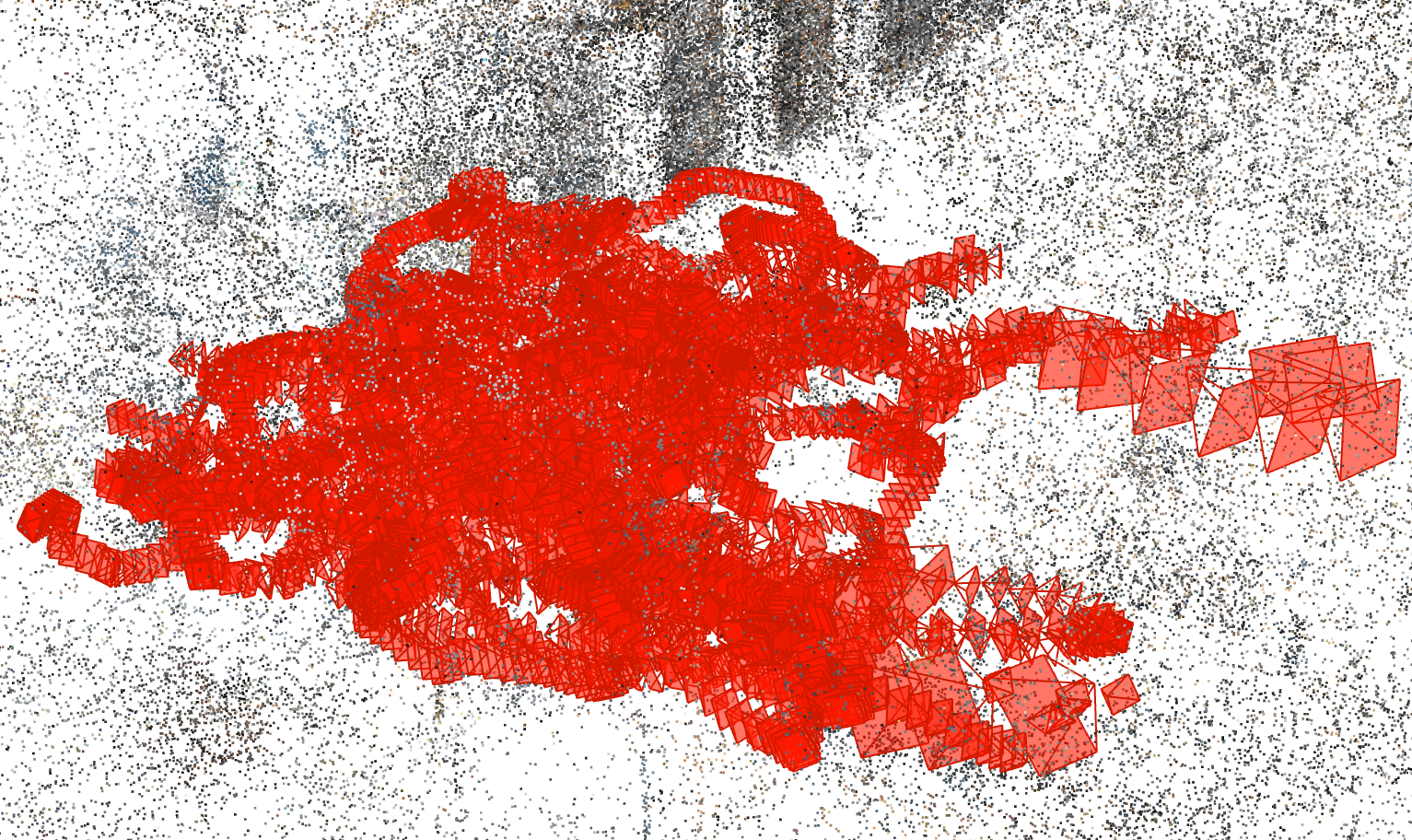}
            \caption{EgoGen synthetic cameras}
    \label{fig:second}
        \end{subfigure}
        \vskip\baselineskip
        \begin{subfigure}[b]{0.203\textwidth}   
            \centering 
            \includegraphics[width=\textwidth]{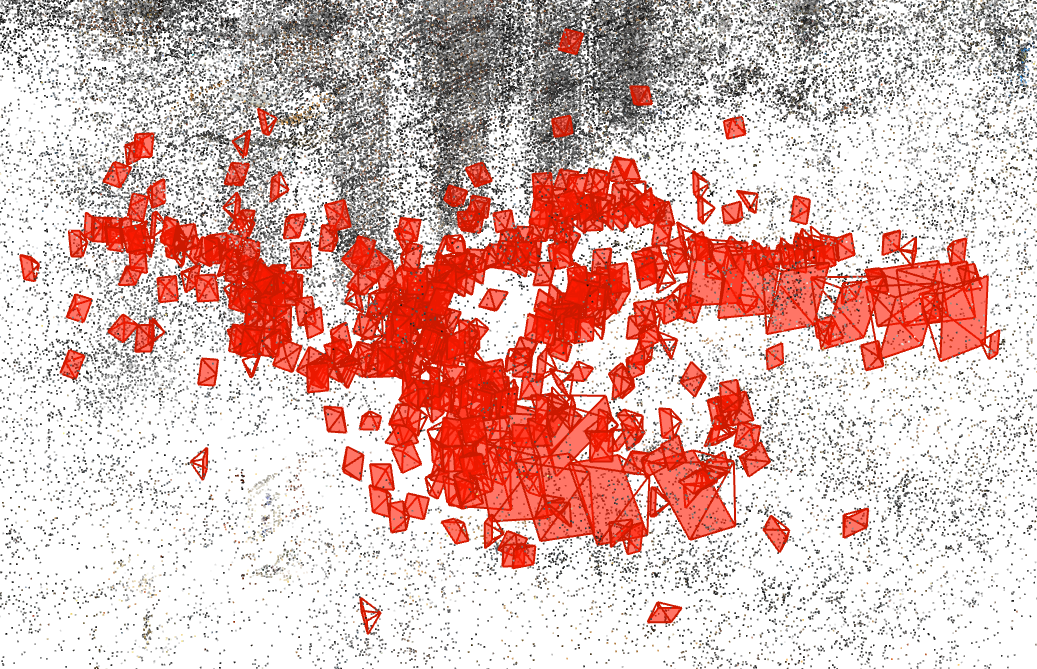}
            \caption{Perturbing existing cameras}
    \label{fig:third}
        \end{subfigure}
        \begin{subfigure}[b]{0.213\textwidth}   
            \centering 
            \includegraphics[width=\textwidth]{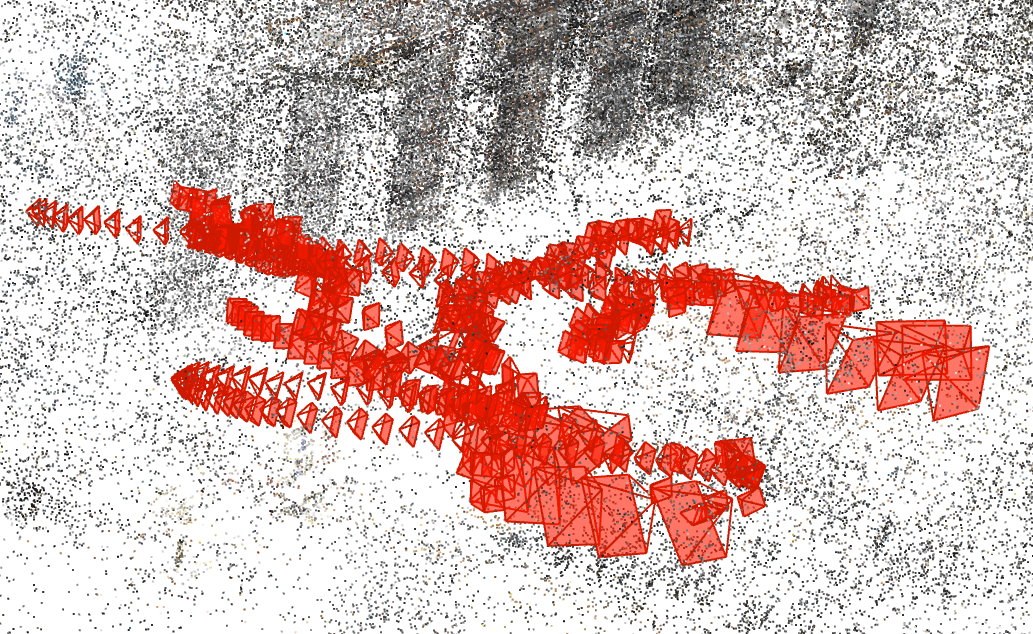}
            \caption{EgoGen same \# of cams as (c)}
    \label{fig:fourth}
        \end{subfigure}
        \caption{EgoGen addresses the issue of sparsity by populating the dataset with synthetic images. In Fig.~\ref{fig:first}, the sparsity of real-world mapping images is apparent, where each red object represents a camera and each colored dot represents a triangulated 3D point. After applying EgoGen, mapping images are more densely distributed, resulting in denser 3D triangulated points, as shown in Fig.~\ref{fig:second}. In Fig.~\ref{fig:third} and Fig.~\ref{fig:fourth}, we augment \ref{fig:first} with the same amount of synthetic images using \cite{ng2021reassessing} and EgoGen respectively. EgoGen generates synthetic data with a similar distribution as human trajectories as illustrated in \ref{fig:fourth}. Results are visualized using Colmap~\cite{Schoenberger2016Structure}. Note that we only visualize a subset of cameras here.\label{fig:mappingmore}}
    \end{figure}

As shown in~\cref{fig:mappingmore}, we can leverage \methodname~to explore the large-scale scene, add synthetic egocentric images into the dataset, and build a more complete Structure-from-Motion (SfM) map (\cref{fig:second}). In our implementation, we randomly set the starting and target locations of virtual humans. Compared with~\cite{ng2021reassessing} that perturbs real-world cameras with random noise (\cref{fig:third}) that may result in unrealistic camera poses, \methodname~can simulate human trajectories and motion (\cref{fig:fourth}).

\begin{figure}[h]
    \centering
    \setlength{\tabcolsep}{0.0130\linewidth}
    \includegraphics[width=1.0\linewidth]{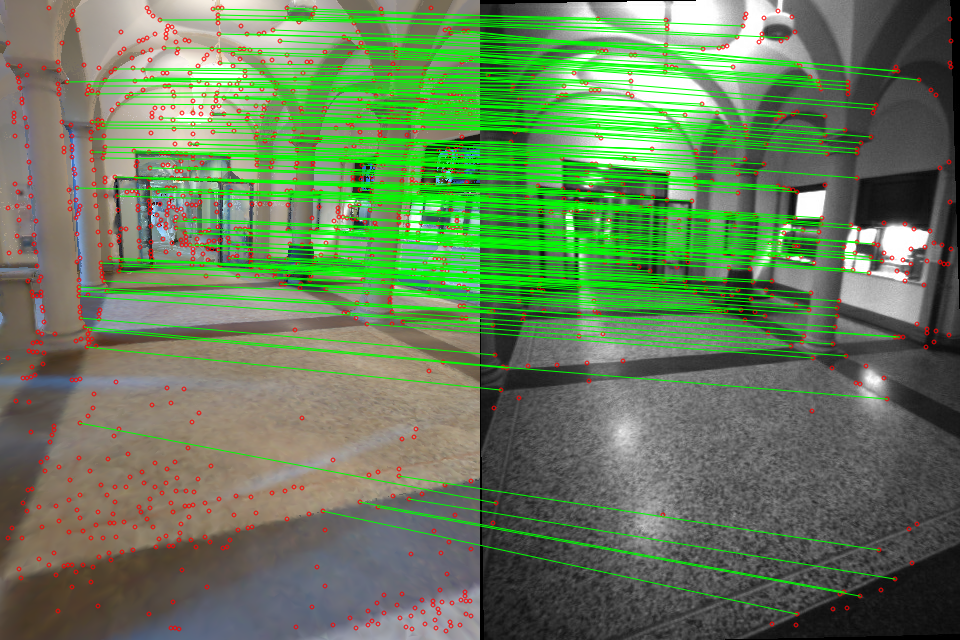}%
    \caption{Feature matching visualization for a render-to-real image pair.
      \label{fig:matching}}
\end{figure}

The efficacy of synthetic data for a task relies on the domain gap between synthetic and real-world images. 
In the SfM pipeline in our experiment, the render-to-real gap can influence the result of the feature extraction. As shown in~\cref{fig:matching} on the left, detected feature points with SuperPoint~\cite{detone2018superpoint} are much noisier in synthetic images due to scene quality, which can make feature matching challenging. In addition, the feature matcher SuperGlue \cite{sarlin2020superglue} exhibits overfitting behaviors: visually similar images are preferred to be matched first, i.e., it tends to match sim-sim and real-real pairs only. As a result, simply adding synthetic images into the real-world dataset will result in no matches between synthetic and real-world images, making it impossible to improve localization recall.

To ensure valid matching between synthetic and real-world mapping images, during the pair selection process using SuperGlue, we force synthetic images to match with real images only.
By implementing this approach, we can achieve a denser SfM map by establishing matches between synthetic and real-world 2D image feature points (see~\cref{fig:matching}) and thereby triangulating more 3D points.

To enhance the localization performance of real-world query images using the augmented SfM map, we enforce matches with both synthetic and real mapping images for all query images. This ensures that real-world query images can be paired with synthetic mapping images, leveraging a denser SfM map and enhancing localization recall.

\subsection{Egocentric Camera Tracking}
\label{sec:s-egoego}
The egocentric camera tracking task is evaluated using the head rotation error, translation error, and pose error that jointly accounts for both rotation and translation.
The head rotation error calculates the Frobenius norm of the difference between the matrix representations of the predicted rotation $R_{pred}$ and the ground truth rotation $R_{gt}$, which is defined as:
\begin{equation}
e_{rotation} = \| R_{pred}R_{gt}^{-1} \|_2,
\label{equ:rot_error}
\end{equation}
The head translation error is computed as the mean Euclidean distance of two sequences of head translations. The results are reported in the unit of millimeter.

The head pose error calculates the Frobenius norm of the difference between the transformation matrix of the predicted head pose and ground truth head pose, which is given by:
\begin{equation}
e_{pose} = \| T_{pred}T_{gt}^{-1} \|_2,
\label{equ:rot_full_error}
\end{equation}
\subsection{Human Mesh Recovery from Egocentric Views}
\label{sec:shmr}

We simulate the data collection process of Egobody~\cite{Zhang:ECCV:2022} and let two virtual humans walk in the scanned scene meshes from Egobody. We randomly sample {\it gender}, {\it body shape}, and {\it initial body pose} and synthesize human motions with our proposed generative human motion model to increase data diversity. 

The egocentric camera is attached to both humans and we render the interactee from the camera wearer's egocentric view. Camera intrinsic is set similarly to the real-world camera. For depth data generation, we omit the clothing because the simulated depth sensor noise will remove detail. For RGB data generation, to further increase data diversity and close the sim-real gap, we randomly sample body texture and 3D textured clothing meshes from BEDLAM~\cite{Black:CVPR:2023} and perform automated clothing simulation (\cref{sec:sclothing-sim}) given arbitrary synthesized human motion sequences from our generative human motion model. In addition, we adopt random lighting in the rendering. In total, we synthesized 105k depth images and 300k RGB images with diverse body shapes, poses, skin textures, and clothing, along with ground-truth SMPL-X annotations. We will release both of our synthetic datasets as a complement to Egobody.

\paragraph{Qualitative results.}

\begin{figure}[h]
    \centering
        \begin{subfigure}[b]{0.45\textwidth}
            \centering
            \includegraphics[width=\textwidth]{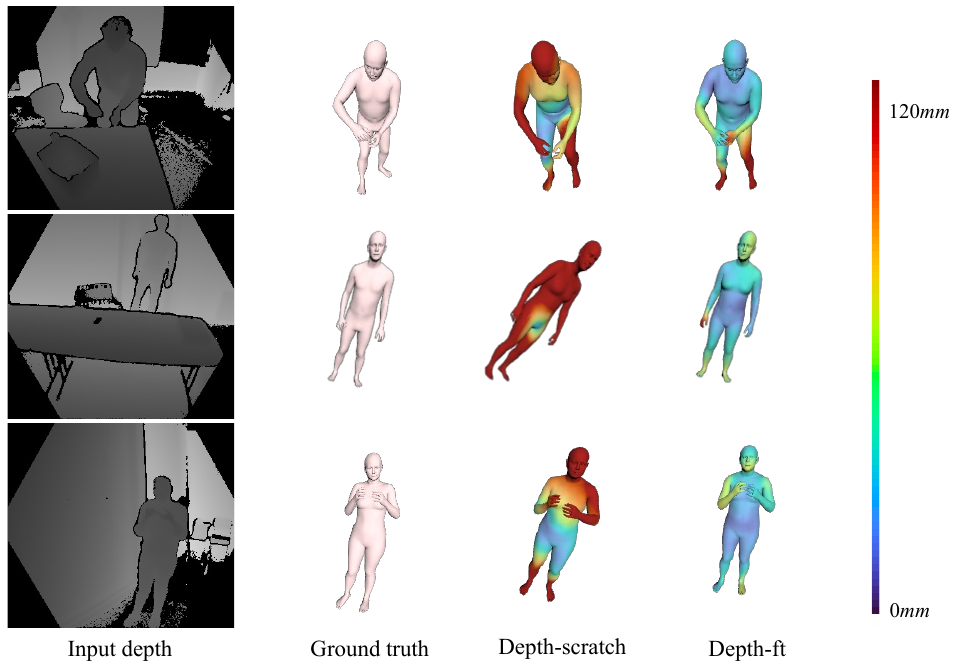}
            \caption{Human Mesh Recovery from Depth Images}
            \label{fig:depth-prohmr}
        \end{subfigure}
        \vskip\baselineskip
        \begin{subfigure}[b]{0.45\textwidth}   
            \centering 
            \includegraphics[width=\textwidth]{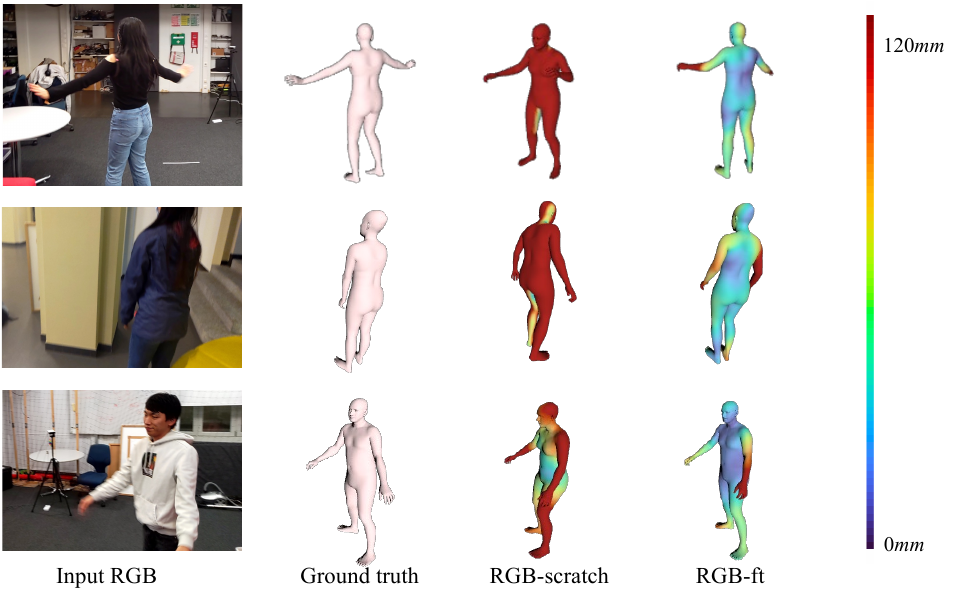}
            \caption{Human Mesh Recovery from RGB Images}
            \label{fig:rgb-prohmr}
        \end{subfigure}
    \caption{Qualitative results of HMR on EgoBody test set. The body mesh color of the last two columns denotes the per-vertex error between the predicted body and the ground truth.}
\label{fig:exp-egobody}
\end{figure}

We visualize our qualitative results for HMR from depth in~\cref{fig:depth-prohmr} and HMR from RGB in~\cref{fig:rgb-prohmr} on real-world test data. With large-scale synthetic data from \methodname, we can compensate for the lack of real-world data and improve the performance of current models. ``*-scratch'' denotes models trained only with limited real-world data. ``*-ft'' denotes models pretrained with our large-scale synthetic data and then finetuned with real-world data.

\begin{figure}[h]
    \centering
        \begin{subfigure}[b]{0.45\textwidth}
            \centering
            \includegraphics[width=\textwidth]{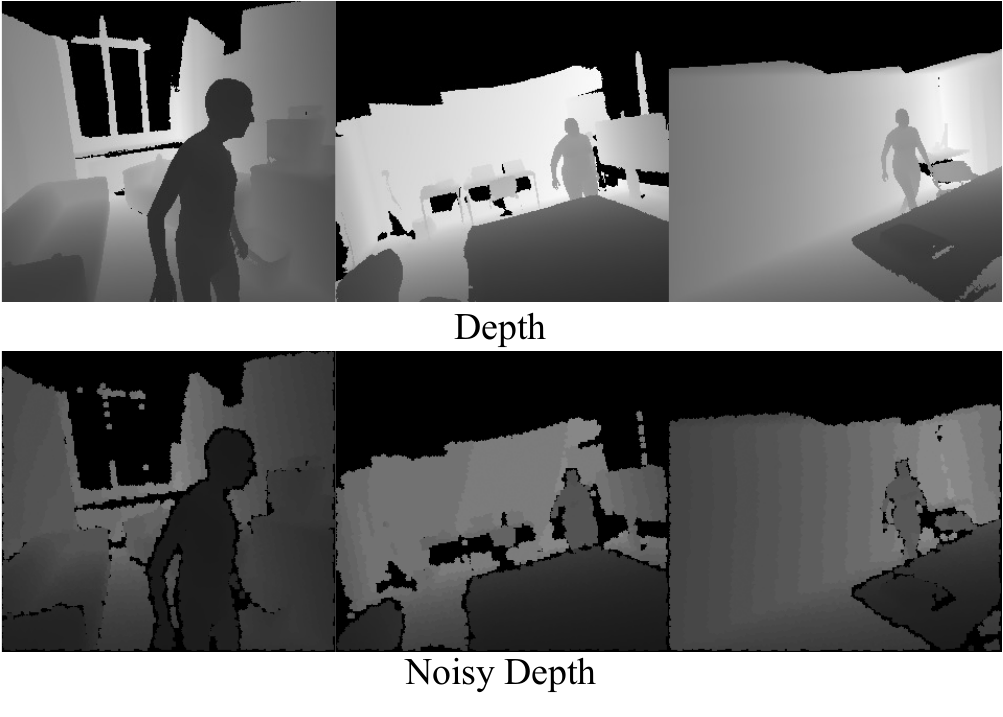}
            \caption{Our synthetic depth images.}
            \label{fig:depth-sample}
        \end{subfigure}
        \vskip\baselineskip
        \begin{subfigure}[b]{0.45\textwidth}   
            \centering 
            \includegraphics[width=\textwidth]{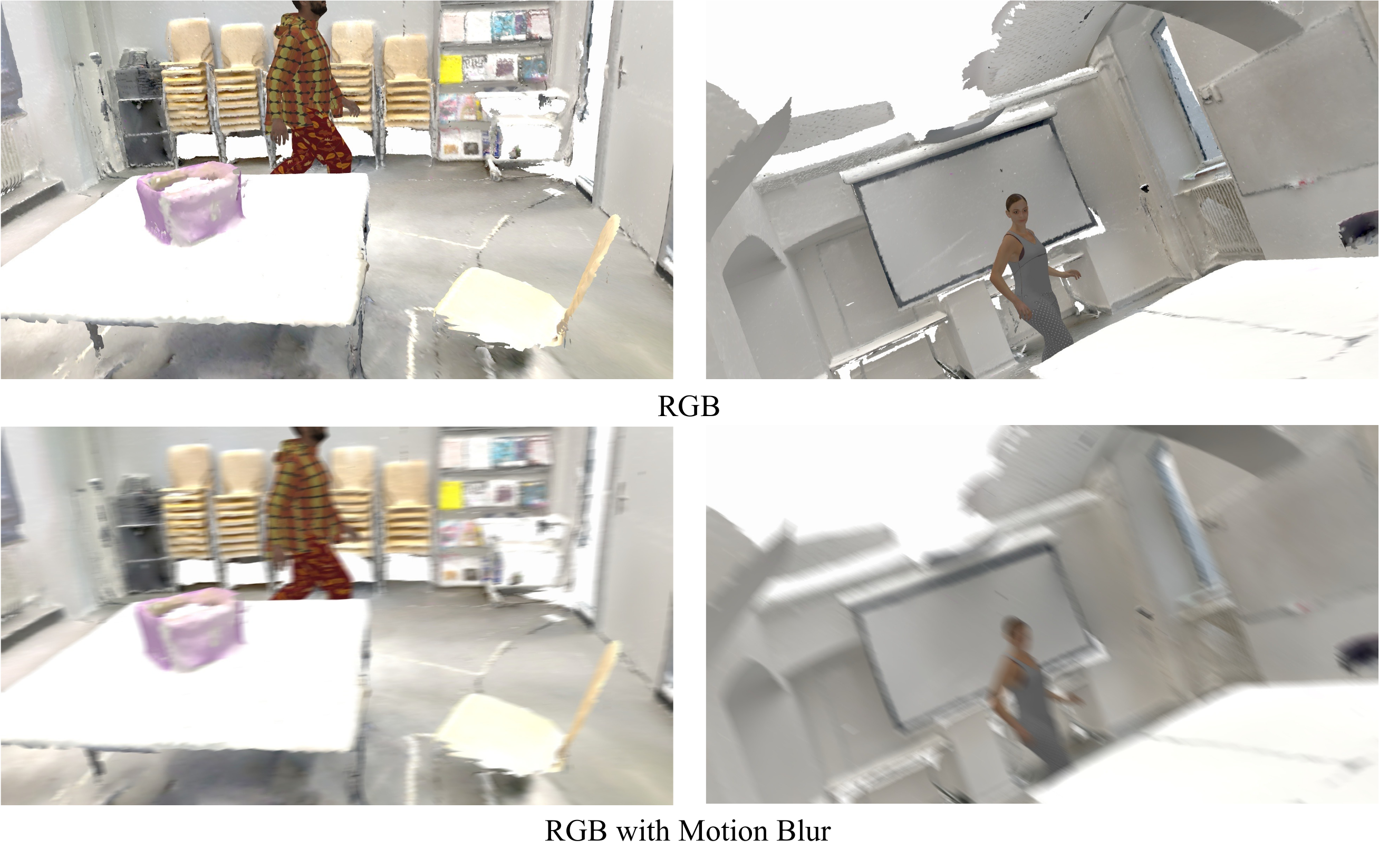}
            \caption{Our synthetic RGB images.}
            \label{fig:rgb-sample}
        \end{subfigure}
    \caption{Synthetic data samples from \methodname.}
\label{fig:egobody-sample}
\end{figure}

\paragraph{Synthetic Data Samples.}
We show some examples of synthetic data from \methodname~in~\cref{fig:egobody-sample}.

\paragraph{Synthetic Dataset Statistics.}

\begin{figure}[h]
    \centering
        \begin{subfigure}[b]{0.45\textwidth}
            \centering
            \includegraphics[width=0.85\textwidth]{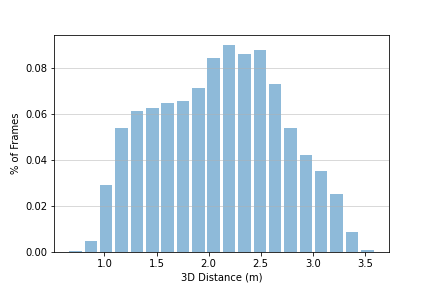}
            \caption{3d Distance on depth images.}
            \label{fig:depth_dist}
        \end{subfigure}
        \begin{subfigure}[b]{0.45\textwidth}   
            \centering 
            \includegraphics[width=0.85\textwidth]{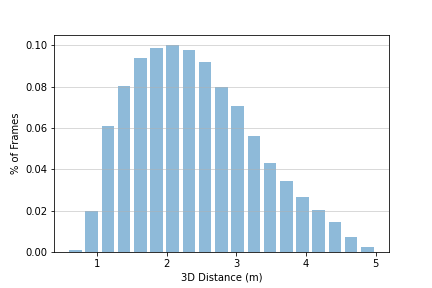}
            \caption{3d Distance on RGB images.}
            \label{fig:rgb-dist}
        \end{subfigure}
    \caption{Interaction Distance of synthetic samples from \methodname.}
\label{fig:egogen_dist}
\vspace{-2mm}
\end{figure}

The generated depth dataset consists of 105000 depth images with 47107 male and 57893 female images. The generated RGB dataset consists of 301073 depth images with 147862 male and 153211 female images. Both datasets cover a large range of indoor interaction distances ranging from $0.60m$ to $5.02m$. \cref{fig:depth_dist} shows the distribution of the interaction distance of the depth dataset and~\cref{fig:rgb-dist} shows the distribution of the RGB dataset.

Additionally, we consider two types of ``invisibility'' of the joints: frame-wise invisibility and joint-wise invisibility ratio. The frame-wise invisibility ratio calculates the percentage of joints that are not on the image plane among all body joints. The joint-wise invisibility ratio calculates the ratio of frames when the joint is
out of the image plane among all frames.

\begin{figure}[h]
    \centering
        \begin{subfigure}[b]{0.45\textwidth}
            \centering
            \includegraphics[width=0.85\textwidth]{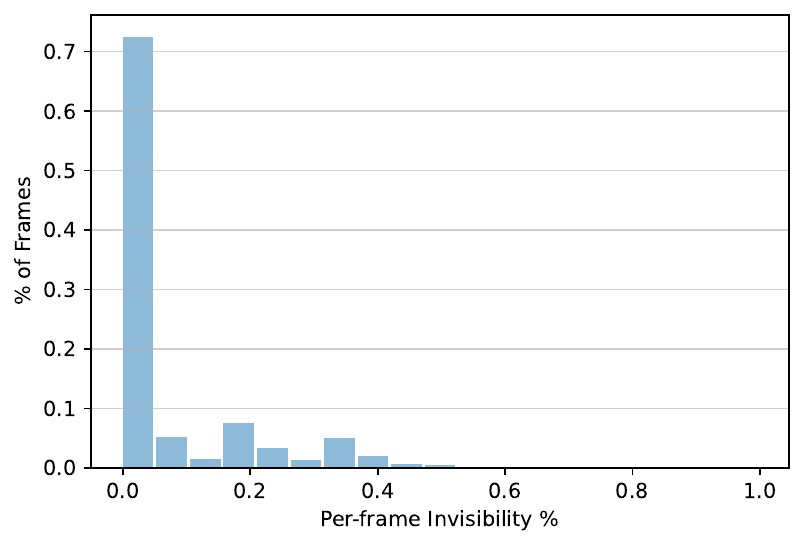}
            \caption{Frame-wise invisibility on depth images.}
            \label{fig:depth_per_frame}
        \end{subfigure}
        \begin{subfigure}[b]{0.45\textwidth}   
            \centering 
            \includegraphics[width=0.85\textwidth]{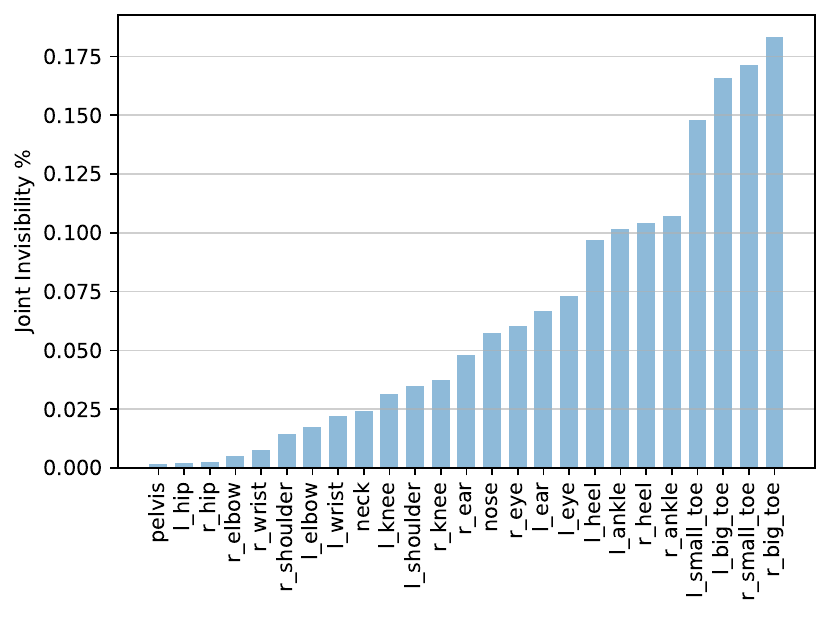}
            \caption{Joint-wise invisibility on depth images.}
            \label{fig:depth_joint_invisibility}
        \end{subfigure}
        \vskip\baselineskip
        \begin{subfigure}[b]{0.45\textwidth}
            \centering
            \includegraphics[width=0.85\textwidth]{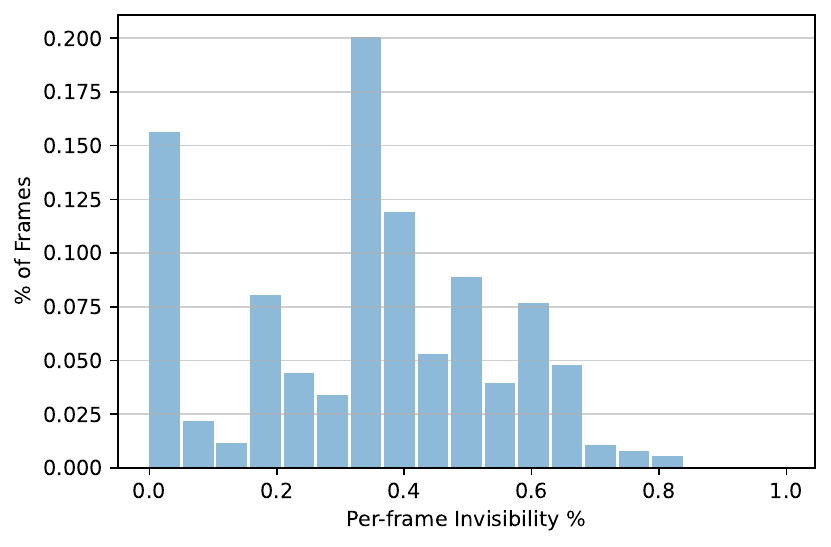}
            \caption{Frame-wise invisibility on RGB images.}
            \label{fig:rgb_per_frame}
        \end{subfigure}
        \begin{subfigure}[b]{0.45\textwidth}   
            \centering 
            \includegraphics[width=0.85\textwidth]{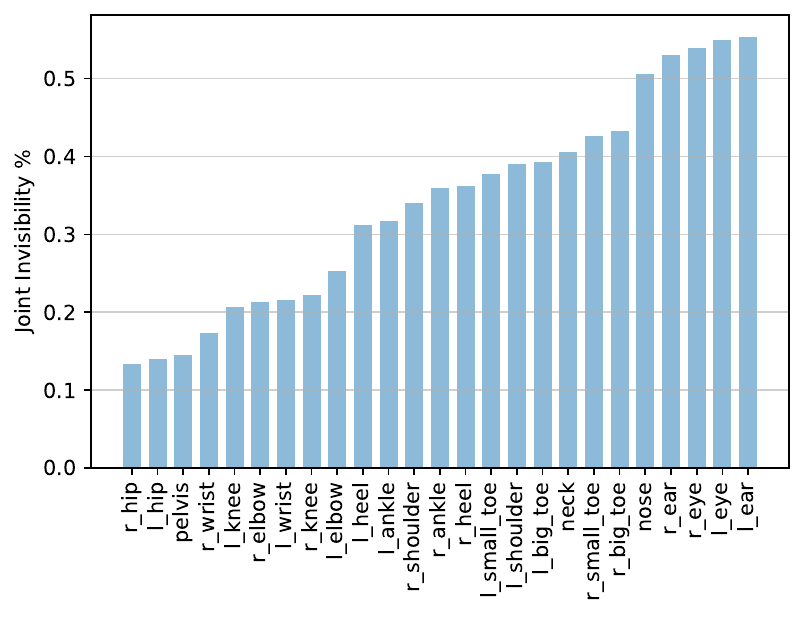}
            \caption{Joint-wise invisibility on RGB images.}
            \label{fig:rgb_joint_invisibility}
        \end{subfigure}
    \caption{Invisibility Statistics.}
\label{fig:depth_invis}
\end{figure}

An analysis of the invisibility distribution of the depth dataset and the RGB depth distribution can be seen in~\cref{fig:depth_invis}. Due to the different camera intrinsic of depth and RGB sensor, the invisibility distribution is different. From~\cref{fig:depth_per_frame}, we can see that over $79\%$ of the depth frame contains more than $90\%$ joints. This means most depth images contain the full body. While from~\cref{fig:rgb_per_frame} we can see that the RGB dataset yields higher invisibility.  The detailed invisibility of the joints is shown in~\cref{fig:rgb_joint_invisibility}. It illustrates that even though RGB images have a higher invisibility, the most frequently missing joints are the upper or lower part of people (eyes and toes). In more than $85\%$ of the images, the pelvis joint can be found. %

\paragraph{Training Details.}

We use data augmentation on the training dataset besides adding the motion blur. These methods include using different kinds of image compression, brightness and contrast modification, noise addition, gamma, hue and saturation modification, conversion to grayscale, and downscaling techniques. During training, we set the batch size to 64 for the training on the depth dataset and 128 for the training on the RGB dataset. We use the AdamW Optimizer in the training process. 

\paragraph{Impact of Training Schemes.}

\begin{table}[t]
 \caption{Training scheme comparison of HMR. ``-scratch'': training with limited real-world data only. ``-ft'': transfer learning. ``-Mixed training'': training with mixed real and synthetic data together.}
\centering
\begin{adjustbox}{width=0.45\textwidth}
 \begin{tabular}{||c c c c c||} 
 \hline
  (All units are \textit{mm}) & G-MPJPE $\downarrow$ & MPJPE $\downarrow$ & PA-MPJPE $\downarrow$ & V2V $\downarrow$ \\ 
 \hline\hline
 Depth-scratch  & 117.7 & 82.2 & 54.1 & 100.6\\ 
 Depth-ft  &\textbf{90.7}   &\textbf{65.2}  &\textbf{47.3}  &\textbf{81.0}  \\
 Depth-Mixed training  & 99.8     & 72.2   & 51.5    & 90.7    \\
 \hline\hline
 RGB-scratch & - & 90.7 & 59.9 & 102.1\\
 RGB-ft & - & \textbf{85.3} & \textbf{56.2} & \textbf{97.2} \\
 RGB-Mixed training  &  -   &85.5    &57.3    &98.2    \\
 \hline
 \end{tabular}
\end{adjustbox}
 \label{tab:s-hmr}
 \vspace{-2mm}
\end{table}

We adopted transfer learning to improve generalization by transferring knowledge from pretraining on synthetic data and refining features through real-world data finetuning.
Mixed training, as shown in \cref{tab:s-hmr}, is less effective than ``-ft'', but still better than training with limited real data only.

\end{document}